\documentclass{article} %

\usepackage{iclr2027_conference,times}  %

\usepackage{amsmath,amsfonts,bm}

\def\eqref#1{equation~\ref{#1}}

\def\1{\bm{1}}

\DeclareMathAlphabet{\mathsfit}{\encodingdefault}{\sfdefault}{m}{sl}
\SetMathAlphabet{\mathsfit}{bold}{\encodingdefault}{\sfdefault}{bx}{n}

\usepackage{graphicx}
\usepackage{amsmath}
\usepackage{amssymb}
\usepackage{booktabs}
\usepackage{subcaption}
\usepackage{multirow}
\usepackage{xcolor}
\usepackage[utf8]{inputenc}
\usepackage{xspace}
\usepackage{tikz}
\usepackage{titletoc}
\usepackage{wrapfig}
\usepackage{capt-of}

\newcommand{\method}{{\textsc{DyCraft}}\xspace}
\newcommand{\benchmark}{{\textsc{DyBench}}\xspace}

\usepackage[table]{xcolor}

\definecolor{studyheader}{HTML}{F1F3F5}
\definecolor{oursgreen}{HTML}{E2F3E8}
\definecolor{ourstext}{HTML}{176B3A}

\usepackage[most]{tcolorbox}
\usepackage{xcolor}

\definecolor{promptbg}{HTML}{F7F8FA}
\definecolor{promptborder}{HTML}{CBD5E1}
\definecolor{promptaccent}{HTML}{4F6F9F}
\definecolor{prompttitle}{HTML}{334155}

\newtcolorbox{promptbox}[1][]{
    enhanced,
    breakable,
    colback=promptbg,
    colframe=promptborder,
    boxrule=0.45pt,
    leftrule=2.5pt,
    left=7pt,
    right=7pt,
    top=6pt,
    bottom=6pt,
    arc=1.5pt,
    before skip=7pt,
    after skip=7pt,
    fontupper=\small,
    coltitle=prompttitle,
    fonttitle=\bfseries\small,
    title=#1,
}

\definecolor{promptBlue}{HTML}{5277A3}
\definecolor{promptBg}{HTML}{F7F8FA}
\definecolor{promptTitle}{HTML}{27384D}
\definecolor{promptLabel}{HTML}{5277A3}

\newcounter{dybenchprompt}

\newenvironment{dybenchprompt}[1]
{
  \refstepcounter{dybenchprompt}
  \begin{tcolorbox}[
    enhanced,
    breakable,
    colback=promptBg,
    colframe=promptBg,
    boxrule=0pt,
    borderline west={2.2pt}{0pt}{promptBlue},
    sharp corners,
    left=9pt,
    right=9pt,
    top=8pt,
    bottom=8pt,
    before skip=8pt,
    after skip=8pt,
  ]
  {\sffamily\fontsize{7.5}{8.5}\selectfont
   \bfseries\color{promptLabel}
   PROMPT \ifnum\value{dybenchprompt}<10 0\fi\arabic{dybenchprompt}}
   
  \vspace{2pt}
  
  {\bfseries\color{promptTitle}#1}
  
  \vspace{5pt}
  
  \small
  \setlength{\parindent}{0pt}
  \setlength{\parskip}{0pt}
}
{
  \end{tcolorbox}
}

\definecolor{citeblue}{HTML}{5B9BD5}
\usepackage[breaklinks,colorlinks,citecolor=citeblue]{hyperref}
\usepackage[capitalize]{cleveref}
\crefname{section}{Sec.}{Secs.}
\crefname{table}{Tab.}{Tables}
\crefname{figure}{Fig.}{Figures}
\crefname{tcb@cnt@example}{Ex.}{Examples}
\crefname{tcb@cnt@definition}{Def.}{Definitions}
\crefname{tcb@cnt@theorem}{Thm.}{Theorems}
\crefname{tcb@cnt@corollary}{Cor.}{Corollaries}
\usepackage{url}

\newcommand{\phbox}[3][0.618]{%
  \begin{tikzpicture}
    \node[draw=gray!60, dashed, line width=0.6pt, fill=gray!8,
          minimum width=#2, minimum height={#1*#2},
          inner sep=0pt, align=center, text=gray!45!black,
          font=\small\itshape, text width={0.9*#2}]
         {#3};
  \end{tikzpicture}%
}

\newcommand{\figorph}[3][0.618]{%
  \IfFileExists{#2}{\includegraphics[width=#3]{#2}}%
                   {\phbox[#1]{#3}{\ttfamily\footnotesize #2\\(missing)}}%
}
\newcommand{\qualrow}[2]{{\footnotesize\textbf{#2}}\\[-1pt]\includegraphics[width=\linewidth]{#1}}

\title{DyCraft: An Agentic System for Long-Horizon 4D Scene Generation}

\author{Jeff Tan$^{*,1}$, Hang Dai$^{*,1}$, Zeqian Long$^{*,1}$, Yanzhe Lyu$^1$, \\
\bf R. Kenny Jones$^1$, Shangzhe Wu$^2$, Chen Geng$^{\dagger,1}$, Jiajun Wu$^{\dagger,1}$ \\
$^1$Stanford University ~~~ $^2$University of Cambridge
\textcolor{white}{\thanks{Co-first author. Work done when H. Dai was a visiting student at Stanford. $^\dagger$Equal advising.\\ Correspondence to \texttt{jefftan1@stanford.edu}. Project page: \href{https://dy-craft.github.io}{https://dy-craft.github.io}}
}}

\iclrfinalcopy %

\begin{document}

\maketitle

\vspace{-2\baselineskip}
\begin{center}
  \includegraphics[width=0.93\linewidth]{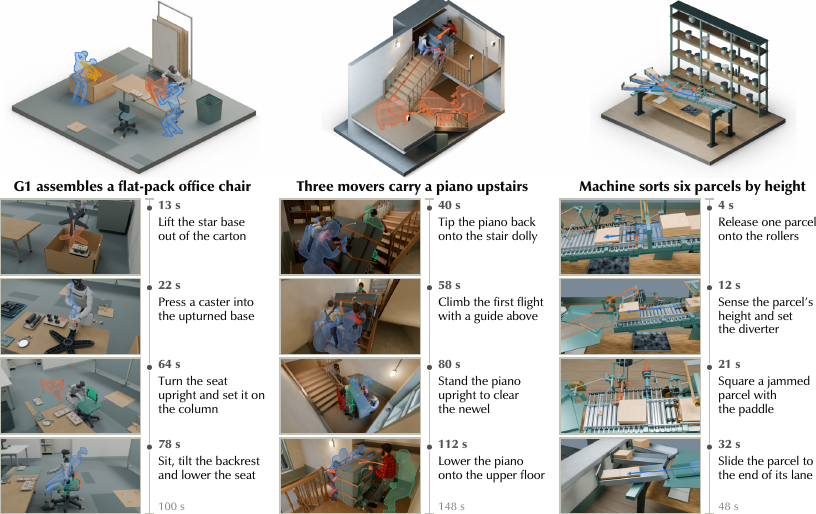}
  \captionsetup{hypcap=false}
  \captionof{figure}{\textbf{\method is an end-to-end agentic system for generating long-horizon animated 3D scenes from text prompts.} Our system allows GPT-6-Astra coding agents to write minute-long 3D animations in Blender, depicting humanoid robots, human characters, and physical mechanisms. 3D moments from the authored films are shown above. Project page: \href{https://dy-craft.github.io}{https://dy-craft.github.io}}
  \label{fig:teaser}
\end{center}

\begin{abstract}

While frontier coding agents can generate high-quality 3D scenes and articulated assets, they struggle to create coherent long-horizon animations.
When equipped with general-purpose harnesses, they frequently produce physically implausible outputs, generating objects that interpenetrate, lack physical support, or move unnaturally.
Such \textit{bare agents} struggle to repair these errors because they default to scripting the full animation in one pass, making it difficult to verify their work and disentangle complex spatiotemporal constraints.
Drawing on the practice of human animators, we introduce \textbf{\method}, a system for animated 3D scene generation that breaks animation into steps that coding agents can more reliably solve. 
\method first expands the user's prompt into a detailed plan and builds a static scene. 
It then creates a 3D storyboard of key animation poses, where each frame is a static 3D scene that the agent can build and debug in isolation.
Storyboarding decomposes the problem into pairs of verified start and end states, allowing the agent to author local motion between them with a collection of motion authoring tools.
We evaluate \method on 24 long-horizon, multi-shot animation tasks covering humanoid robots, human and animal characters, and physical mechanisms.
Compared with a native GPT-6 agent in Codex and prior agentic baselines, \method substantially reduces physical violations and generates higher-quality animated 3D scenes that are more often preferred by human raters.

\end{abstract}

\section{Introduction}
\label{sec:intro}

Animated 3D content is labor-intensive to produce, yet fundamental to filmmaking, gaming, and robot learning. Although video generative models can produce highly realistic footage, their pixel-space outputs are difficult to precisely control and often exhibit geometric, temporal, and physical inconsistencies \citep{cai2024generative, sun2025dimensionx, fang2025worldreel, song2026taming, gu2026phyworldbench}. 
Meanwhile, learned 4D generation methods \citep{wu2025animateanymesh, yin2026sculpt4d} suffer from scarce training data, limiting such approaches to short object-centric animation sequences.

\begin{figure}[t]
  \centering
  \includegraphics[width=\linewidth]{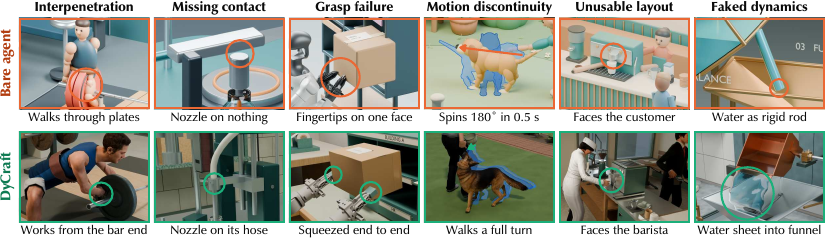}
  \caption{\textbf{Failure modes of bare agents.}
  The top row shows GPT-6-Astra with Codex producing interpenetration, missing contact, grasp failure, motion discontinuity, unusable layout, and faked dynamics.
  The bottom row shows corresponding results produced by \method with GPT-6-Astra.}
  \label{fig:failure}
  \vspace{-0.3cm}
\end{figure}

A compelling alternative, given recent advances in coding agents, is to represent animated content as explicit code for 3D creation tools such as Blender \citep{blender}. Representations based on code are controllable and persistent, naturally enabling downstream tasks that require geometric consistency or precise physical interaction. Although agentic systems \citep{hu2024scenecraft, lu2025ll3m, zhou2026articraft, pfaff2026scenesmith} and general-purpose coding agents can generate highly detailed 3D scenes and articulated assets, agentic generation of 4D (i.e. 3D+time) content has received comparatively little attention. Most existing systems generate short sequences often based on simulation \citep{yin2026vision,zhang2026code2worlds, wang2026coding, liu2026simworlds}, animate a single articulated mesh \citep{zou2026artimo}, or compose pre-built characters and environments into cutscenes without substantial dynamic motion \citep{xu2025filmagent, he2026cutscene}. Even frontier coding agents \citep{openai2026astra} exhibit systematic failures during long-horizon animation (Fig. \ref{fig:failure}), including
frequent interpenetration, 
objects that lack physical support, 
and implausible motions.

We attribute these failures mainly to the agent's workflow rather than its coding abilities. Animation quality depends on subtle temporal and spatial relations across many frames, making animations difficult to author and repair locally. Traditional animation practice addresses this by planning and reviewing key poses before producing intermediate motion, commonly called \emph{pose-to-pose animation} \citep{thomas1981disney,lasseter1998principles}. Applied to coding agents, this practice has two advantages: (1) it reduces long-horizon animation authoring to a sequence of static 3D scenes, drawing on agents' strong 3D modeling ability; and (2) it allows subsequent motion to be authored between 3D states whose physical validity was already verified. We observe that general coding agents lack this workflow and instead author full animations in a single pass without systematic verification.

In this work, we present \method, which equips coding agents with an animation workflow, domain-specific authoring guidance, and systematic verification while leaving authoring autonomy to the agent. \method contains three components: (1) a planning stage that expands the user's prompt into a detailed scene and motion plan; (2) an authoring stage that plans key states as a 3D storyboard, renders and reviews those states, then authors continuous motion between the reviewed states;
and (3) a verification stage that validates per-frame geometry and motion for interpenetration, contact, and physical support. \method also provides domain guidance and solvers for locomotion and grasping, as well as a structured CAD-based representation for object and scene authoring.

We evaluate \method on a benchmark of 24 long-horizon multi-shot animation briefs covering humanoid robots, human and animal characters, and physical mechanisms. Compared to general coding agents and existing specialized agentic systems using the same underlying model, \method substantially reduces physical violations (Sec. \ref{sec:eval_physics}) and is preferred in a user study (Sec. \ref{sec:eval_user_study}). %

\section{Related Work}
\label{sec:related}

\paragraph{4D generation.} Automatically generating 4D scenes with long-horizon animation is a long-standing challenge in computer graphics. We can view this process as a joint planning and optimization problem over several interdependent components, including objects, rigging, motions, and simulation. Prior learning-based methods handle these components one by one, often for a specific category: generating articulated objects~\citep{chen2024urdformer,jiang2022ditto,kerr2024robot,le2024articulate,lei2023nap,liu2023paris,liu2024singapo,mandi2024real2code,noguchi2022watch,xia2025drawer,xu2025gaussianproperty,li2026art}, rigging objects~\citep{zhang2026rigmo,baran2007automatic,deng2025anymate,he2025category,morig,rignet,xu2019rig}, predicting simulation parameters~\citep{chen2025physgen3d,dagli2025vomp,li2023pac,li2025wonderplay,lin2024phys4dgen,mittal2025uniphy,zhang2024physdreamer}, and synthesizing object motions~\citep{bahmani20244d,Chen_2025_ICCV,chen2026motion3to4,gao2024gaussianflow,geng2025birth,lyu2026choreographing,ma20254d,ren2023dreamgaussian4d,ren2024l4gm,uzolas2025motiondreamer,wang2024vidu4d,xie2024sv4d,yao2025sv4d,zhao2023animate124}. Generating 4D scenes end-to-end was once infeasible, because such training data is scarce. However, agentic systems now let us treat the problem as a whole: we use LLM-powered coding agents to orchestrate, author, and verify across stages, and to co-design the components jointly. This new treatment leads to an end-to-end system that generates long-horizon 4D scenes directly from text prompts.

\paragraph{Agentic creation of graphics assets.} The emergent spatial understanding of frontier models~\citep{openai2026astra, zeng2026glm} enables agentic systems to author visual content by writing code. Prior systems demonstrate this on individual asset types, creating articulated objects~\citep{zhou2026articraft}, 3D scenes~\citep{hu2024scenecraft, pfaff2026scenesmith, wang2026scenecode, guo2026worldclaw, gu2025blendergym,huang2024blenderalchemy,kulits2024re,ling2026scenethesis,sun20253d,yin2026vision,zhang2025scene}, and motion graphics~\citep{ma2025mover, liu2025logomotion, shirakawa2025mg, silva2026training, gunturu2025mapstory}. We extend this ability to long-horizon 4D scenes that contain both complex character motion and object interactions. The closest works to ours, VIGA~\citep{yin2026vision} and SimWorlds~\citep{liu2026simworlds}, generate short 4D scenes through simulation, without explicit kinematic motion. In contrast, our system generates faithful, minute-long 4D scenes in which characters move under their own control and objects respond to physics.

\section{Methodology}
\label{sec:method}

\begin{figure}[t]
  \centering
  \includegraphics[width=\linewidth]{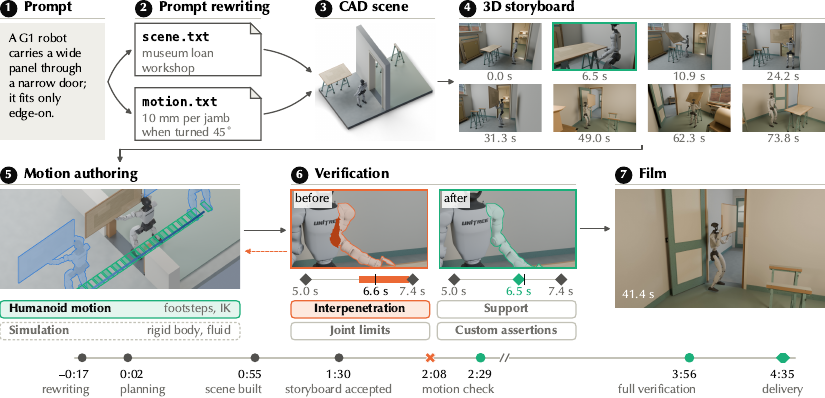}
  \caption{\textbf{Overview of \method on an example.} A humanoid robot carries a wide panel through a narrow doorway.  Following pose-to-pose animation, \method breaks authoring into steps that the agent can check individually. The sparse user prompt (1) is first expanded into detailed scene and motion briefs (2), which the agent uses to build the scene in CAD (3). The agent then plans key poses and object states as a 3D storyboard (4). After inspecting and verifying these states, the agent authors motion connecting them with domain-specific tools (5). Verifiers check every frame (6), catching errors in the storyboard states and motions: here, the arm passes through the torso (orange). The agent repeatedly revises the scene and motion before rendering the final film (7).}
  \label{fig:overview}
\end{figure}

\subsection{Harness Overview}
\label{sec:overview}

We introduce \textbf{\method}, an agentic system that generates long-horizon animated 3D scenes entirely through code. \method extends general coding agent harnesses (\textit{e.g.,} Codex, Claude Code) with a custom system prompt and operations for generating, probing, and verifying scenes and motion. Given a user instruction, the agent uses its native tools (\textit{e.g.,} file editing, shell commands, and Python) alongside our custom harness operations to create an editable Blender project, pass scene- and motion-level checks, visually review the complete rendered animation, and deliver the final film.

For capable coding agents, we find general harnesses to have significant advantages over existing specialized agentic 3D/4D harness systems \citep{zhou2026articraft, liu2026simworlds, wang2026scenecode}, which typically prescribe detailed pipeline stages that restrict the agent's flexibility. \method's hybrid approach hands autonomy to the coding agent while providing domain-specific functionality (\textit{e.g.,} generating and verifying 4D motions) through custom operations. This allows the agent to freely execute code and inspect results, preserving the bare agent's strong iteration and verification abilities while the harness can steer the agent towards animation-specific workflows, provide common tools, and help the agent identify errors that would otherwise require costly iteration. 

\subsection{Planning and Scene Design}

\begin{figure}[t]
  \centering
  \includegraphics[width=\linewidth]{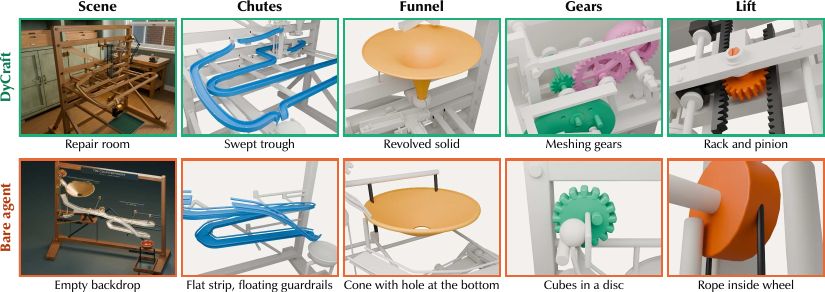}
  \caption{\textbf{Scene authoring with CAD tools.} Scene components authored by \method (top) and by a bare agent
  (bottom) for a marble run (both GPT-6-Astra). CAD operations allow \method to construct
  precise solid geometry, whereas the bare agent approximates the same components with primitives and thickened
  surfaces that intersect or fail to connect.}
  \label{fig:scene_authoring}
\end{figure}

\paragraph{Planning via Prompt Rewriting.} Given a user instruction, our harness first performs two rounds of prompt rewriting: one for the objects and scene, then one for the motion.  
User prompts are often sparse or incomplete, describing a few intended outcomes, while leaving many scene and motion details under-specified. 
We find that bare agents rarely develop these details beyond what the prompt explicitly states. 
We therefore use fresh-context prompt rewriters that read the instruction
and fill in under-specified details by selecting an organizing principle around which to organize the objects, scene, and motion. 
For example, the rewriters are encouraged to consider other activities the scene could support, how objects should be constructed beyond basic primitives, and the spatial relationships and state transitions required for believable motion. They produce detailed scene and motion briefs, which are staged alongside the user instruction and system prompt. Based on these expanded briefs, the agent plans the full action, exact shot sequence, and required objects, motions, and transitions. In Sec. \ref{sec:ablations}, we find that prompt rewriting significantly improves scene and motion detail. The rewriter system prompt is provided in Sec.~\ref{sec:rewriting}.

\paragraph{Scene Construction via CAD.} Given the rewritten prompts and its animation plan, the agent starts to build the static scene and objects. We observe that bare agents build scenes as collections of basic Blender primitives (\textit{e.g.,} cube, cylinder) and use hand-written vertex and face lists for all complex geometry (including extrudes, revolves, and lofts) (Fig.~\ref{fig:scene_authoring}). This limitation requires bare agents to rebuild a CAD-like geometry authoring library anytime they want to author complex geometry. Our harness addresses this limitation with a boundary-representation (B-rep) CAD interface~\citep{build123d, occt}, which allows the agent to create precise solid geometry for animation in Blender. To support articulated object and scene construction, we supplement build123d with additional tools including parts and articulated joints, named coordinate frames, support for posed object instances, and joint limit inspection. 
This operation set is additive, and the agent retains access to native Blender tools for deformable characters, simulated fluids, and other dynamic phenomena that are difficult to model with CAD.
The agent can also choose to incorporate external human, animal, and robot characters from a generic character library (Appendix \ref{sec:supp_characters}).
Overall, as shown in Fig.~\ref{fig:scene_authoring}, CAD authoring improves the agent's ability to write exact high-quality solid geometry needed for animated 3D scene generation. %

\subsection{Motion Authoring with 3D Storyboarding}
\label{sec:storyboarding}

\paragraph{3D Storyboarding.} After constructing the static scene, the agent storyboards the animation before authoring continuous motion.
We observe that bare Blender animation agents tend to script the entire motion in one pass, leaving unrepaired errors across the full animation. Layout and staging problems therefore surface late, intertwined with motion and logic errors, and are difficult to repair without rewriting the sequence as a whole. To address this, we require the agent to author key moments in 3D (\textit{i.e.,} storyboard) before writing continuous motion. The harness renders the storyboard through the film's own cameras and checks each state with 3D verifiers (Sec. \ref{sec:verifiers}). Within each film segment, the agent must pass these verifiers and visually inspect the relevant storyboard frames before it can author motion. 
As rendering and verifying storyboard states is cheap, the agent can quickly iterate on layout, pose, and camera choices, settling them before the motion is authored.
Thus, the storyboard turns otherwise entangled animation problems into static 3D checks: whether a doorway is wide enough, a workspace is within reach, or a camera clearly shows the action.

\begin{figure}[t]
  \centering
  \includegraphics[width=\linewidth]{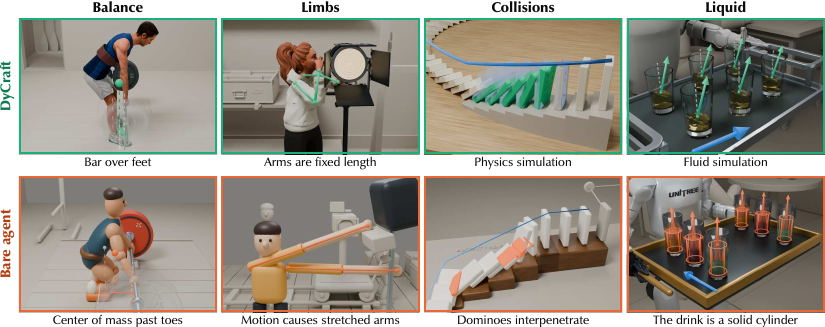}
  \caption{\textbf{Motion authoring tools.} Motion authored by \method (top) and by a bare agent (bottom) for the same
  prompts. Using domain-specific techniques such as physical simulation and inverse kinematics, \method produces
  motion that stays physically plausible, avoids interpenetration, and responds to dynamics.
  The bare agent's hand-scripted motion often violates these constraints. 
  Orange marks
  bare-agent defects and blue arrows indicate time.}
  \label{fig:motion_authoring}
\end{figure}

\paragraph{Motion Authoring Tools.} The verified 3D storyboard anchors the agent's continuous motion authoring: the generated animation must pass through the verified storyboard states.
Although bare agents can in principle use any motion authoring method, they overwhelmingly favor kinematic scripting (Fig.~\ref{fig:motion_authoring}): characters are manually keyframed without physical constraints, objects fall at a constant rate rather than accelerating under gravity, and liquids are represented as static cylinders.

To address these motion authoring limitations, we provide a library of domain-specific skills. 
For example, to address implausible character motion, we introduce a humanoid motion skill that instructs the agent how to author smooth and lifelike motion (\textit{e.g.,} planning footsteps before the body moves, avoiding backwards-bending knees and elbows, etc.). 
Our harness also guides the agent on how to use Blender's native simulation tools, so that physical dynamics such as rigid body collision and liquids are simulated and baked rather than animated by hand. 
Together, this domain-specific authoring guidance greatly improves the animation quality. %

\paragraph{Visual Inspection.} At any time, agents can render selected frames from arbitrary camera viewpoints to inspect the scene and motion. This allows the agent to engage in its own native visual feedback loops (in contrast to VIGA~\citep{yin2026vision}, which prescribes this process). We provide helper functions to assemble grids of rendered images, making it easier to inspect motion over time.

\paragraph{Delegation.} 3D storyboarding decomposes whole-film motion authoring into a series of local subtasks. Likewise, once the scene and its objects are planned, 3D authoring can occur in parallel. To exploit this parallelism potential, agents in our harness can delegate work to object, scene, or motion subagents. The main agent remains responsible for integrating subagent contributions into the film.

\subsection{Verifiers}
\label{sec:verifiers}

\begin{figure}[t]
  \centering
  \includegraphics[width=\textwidth]{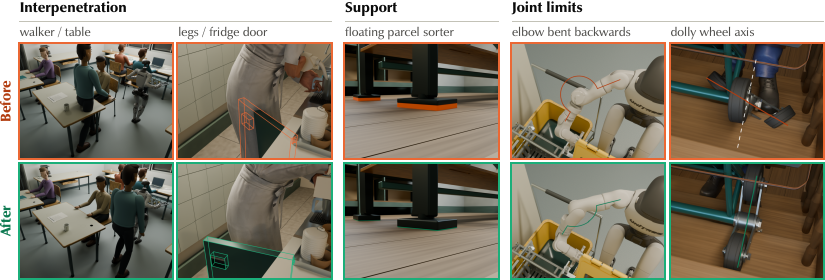}
  \caption{\textbf{Verifiers.} 
  Each column shows a state that fails a verifier (top). 
  After agent repair, the verifier passes (bottom).
  Verifiers for interpenetration, support, and joint limits catch errors that are
  easily missed by visual inspection, such as characters passing through furniture,
  objects floating above their supports, and joints that violate anatomical limits or rotate about a wrong axis.
 }
  \label{fig:verifiers}
\end{figure}

To aid motion authoring, we provide the agent with scene and motion verifiers that perform correctness and physical plausibility checks.
Without external verifiers, bare agents frequently violate basic physical constraints. For instance, they could produce objects that float unsupported, are moved by external ghost forces, or have interpenetrating geometry (Fig.~\ref{fig:verifiers}). 
Even when explicitly prompted to ensure physical plausibility, agents do not reliably verify the full animation. Instead, they usually spot-check a few frames, leaving broader physical layout and timing issues unaddressed.

Our numerical verifiers for objects and scenes require all geometry to pass the same standard set of checks. 
The agent can compile the scene and motion at any time to obtain a report with actionable diagnostics (e.g., penetration depth and separating axis for an intersection between two objects). Any failing verifier blocks clean task completion until the issue is addressed - delivering the film with exceptions is supported only as a fallback. 
Our animation verifiers check whether articulated objects collide with each other over their per-frame poses, whether all objects are physically supported (in a graph connectivity sense), whether articulated objects remain within their authored joint limits, and any custom assertions that the agent declares in its own code. 
Fig.~\ref{fig:verifiers} shows examples of errors detected by these verifiers during generation, together with the corrected states after repair.

\section{Experiments}
\label{sec:experiments}

We evaluate \method across a diverse set of long-horizon animation tasks, aiming to understand whether our harness design improves the animation capabilities of coding agents, as well as which components are responsible for the improved performance. We present qualitative and quantitative results that highlight \method's visual fidelity and diverse capabilities across multiple domains, including humanoid motion, character animation, and physics-based phenomena. For all experiments and baselines, we use GPT-6-Astra within the Codex harness.

\subsection{\benchmark Benchmark}
\label{sec:exp_benchmark}

Existing 4D generation benchmarks~\citep{zhang2026code2worlds, liu2026simworlds, yin2026vision, lyu2026choreographing} do not target long-horizon, multi-shot animation with character motion and object interactions. We therefore introduce \benchmark to stress-test this setting. \benchmark includes 24 text-to-4D generation prompts, split equally into three embodiments: humanoid robots, human/animal characters, and physical phenomena. Representative examples include: a humanoid robot mopping a break room from interior to exit; a barista serving takeaway coffee; a crow removing a peg to open a bird feeder; and an industrial bottling line that fills, caps, and labels beverage bottles. We provide full task briefs in Appendix \ref{sec:supp_dybench_prompts}.

Our baselines include bare agents using the same native coding agent harness and VIGA \citep{yin2026vision}, which uses a custom harness with a code-render-inspect loop. For fair comparison with bare agents, we provide them with a condensed natural language description of our quality standards (Appendix \ref{sec:supp_bare_guidance}), requiring that the final scene be detailed and physically plausible without prescribing our prompt rewriting, planning, authoring, or verification tools. 

\subsection{Qualitative Results}
We show a visual comparison on four prompts from \benchmark in Fig.~\ref{fig:qual}, with results on all 24 prompts in Appendix~\ref{sec:supp_more_results}. \method is able to generate diverse and coherent minute-long animations with multiple types of motion involving humanoid robots, non-human characters, and diverse physical phenomena such as cloth, liquids, and rigid bodies. In comparison, VIGA produces short animations that omit several key events described in the prompts, while the bare agent completes these events in considerably simpler scenes and with lower-quality motion, for example with characters assembled from geometric primitives.

\begin{figure}[p]
\centering
    \begin{minipage}{0.97\linewidth}
  
  \qualrow{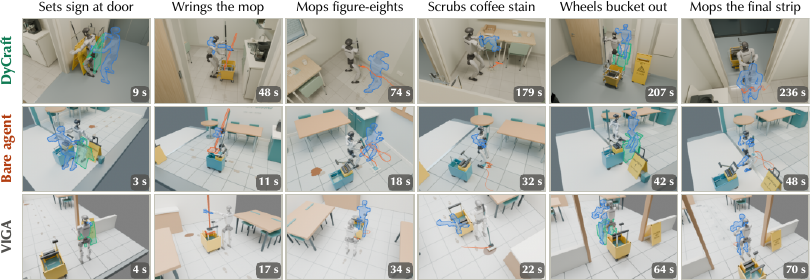}{G1 mops a break room, working backward toward the exit}\\[0pt]
  \qualrow{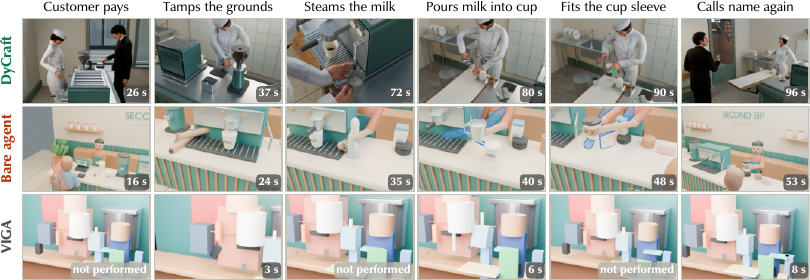}{Barista serves a takeaway coffee}\\[0pt]
  \qualrow{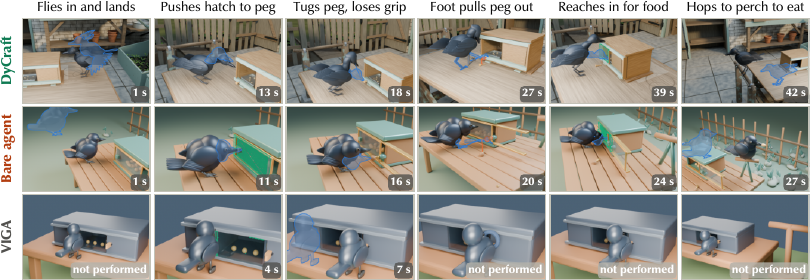}{Crow opens a puzzle feeder}\\[0pt]
  \qualrow{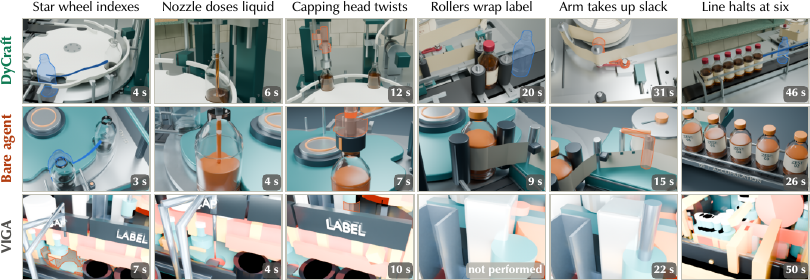}{An automatic line fills, caps, and labels six bottles}
  \end{minipage}
  \vspace{-8pt}
  \caption{\textbf{Qualitative comparison with VIGA and a bare agent.} Please view our video results on the supplemental webpage. Each column corresponds to an event described
  in the prompt and shows the frame at which it occurs in each method's animation; events that a
  method does not produce are marked as not performed. 
  Compared with \method, the bare agent produces simpler scenes with physically implausible motion and VIGA fails to generate several key events.
  Translucent overlays show earlier poses, trails
  indicate fast motion, and the main actor is highlighted in blue.}

  \label{fig:qual}
\end{figure}

\begin{table}[t]
\centering

\begin{minipage}[t]{0.52\textwidth}
\centering
\caption{
\textbf{Human preference evaluation.} Percentage of pairwise judgments in which participants preferred \method generations over each baseline.
}
\label{tab:user_study}
\resizebox{\linewidth}{!}{%
\begin{tabular}{@{}lcc@{}}
\toprule
\textbf{Evaluation Aspect} & \textbf{Ours vs.\ VIGA} & \textbf{Ours vs.\ Bare Agent} \\
\midrule
Overall 4D Realism (\%)    & 92.0 & 75.6 \\
Physical Plausibility (\%) & 88.5 & 70.1 \\
Motion Quality (\%)        & 88.5 & 71.0 \\
Overall Preference (\%)    & 89.7 & 72.3 \\
\bottomrule
\end{tabular}%
}
\end{minipage}\hfill
\begin{minipage}[t]{0.46\textwidth}
\centering
\caption{
\textbf{Physical Plausibility.}
Percentage of frames across all 24 delivered films where a dynamic object is penetrating or floating (Sec. ~\ref{sec:eval_physics}). 
}
\label{tab:physical_eval}
\resizebox{\linewidth}{!}{%
\begin{tabular}{lccc}
\toprule
\textbf{Metric} & \textbf{Bare Agent} & \textbf{VIGA} & \textbf{Ours} \\
\midrule
Penetration Rate (\%) $\downarrow$ & 80.1 & 74.9 & \textbf{13.0} \\
Floating Rate (\%) $\downarrow$    & 15.7 & 26.1 & \textbf{2.7} \\
\bottomrule
\end{tabular}%
}
\end{minipage}

\end{table}

\subsection{Human Preference Evaluation}
\label{sec:eval_user_study}
Recognizing the difficulty of designing reliable automated metrics for dynamic scene generation, we mainly rely on human preference evaluation to assess generation quality. User study participants were presented with generated films from our method alongside a baseline in random order. Participants were then asked three binary-choice questions for each pair: (1) which animation better resembled a real scene captured on camera, (2) which maintained plausible contact and support without interpenetration or floating, and (3) which showed more natural motion across characters, mechanisms, liquids, and other moving elements. We recruited 109 raters on Prolific, each evaluating five video pairs and answering 15 binary-choice questions. As shown in \cref{tab:user_study}, participants preferred our results over VIGA in 89.7\% of judgments and over the bare agent in 72.3\% of judgments. We provide more details in Appendix \ref{sec:supp_user}.

\begin{figure}[t]
  \centering
  \includegraphics[width=\linewidth]{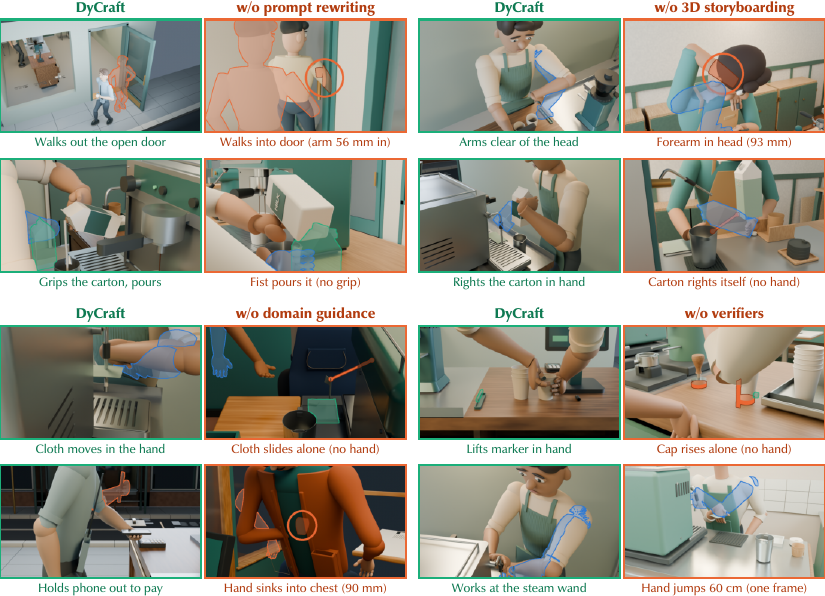}
  \caption{\textbf{Ablation examples on barista making coffee.} For each variant, we show defects over \method on the same action.}
  \label{fig:ablation}
\end{figure}

\subsection{Physical Plausibility Evaluation}
\label{sec:eval_physics}

We further evaluate physical plausibility of the delivered animated scenes using two automated metrics. 
\textbf{(1) Penetration rate} is the fraction of frames with at least one dynamic mesh penetration above a 5mm tolerance, averaged over all 24 delivered films. We exclude persistent construction overlaps in the static scene, directly connected structural parts, liquid or effect meshes, and food that has been fully consumed by a character. The metric includes both self- and cross-object penetration. 
\textbf{(2) Floating rate} is the fraction of dynamic body-frames that lack a valid support path to the ground through physical contact, structural attachment, deformable carriers, or grasping, averaged over all 24 delivered films. We construct a per-frame support graph and determine support by graph reachability from the ground to each moving body.

As shown in \cref{tab:physical_eval}, \method substantially improves physical consistency over both baselines. Across the 24 shared tasks, the penetration rate reduces from 80.1\% for a bare agent and 74.9\% for VIGA to 13.0\% for \method. The remaining failures are primarily between characters and held objects (e.g. a dog's mouth sinking into the newspaper) or layered parts (e.g. stitches on a weight belt). \method further achieves a floating rate of only 2.7\%, compared to 15.7\% and 26.1\% for the bare agent and VIGA, respectively. These results indicate that \method produces animations with substantially fewer geometric interpenetrations and unsupported moving objects.

\subsection{Ablation Studies}
\label{sec:ablations}
We evaluate four method variants in Fig. \ref{fig:ablation} by removing one component at a time: a) w/o planning via prompt rewriting, b) w/o 3D storyboarding, c) w/o domain-specific motion authoring tools, and d) w/o external scene and motion verifiers. Without our full system, all ablation arms exhibit planning or physical plausibility errors, while our complete system has improved performance.

\section{Conclusion}
\label{sec:conclusion}

We introduce \method, an agentic system for generating long-horizon animations entirely as code. By augmenting general-purpose coding agents with domain-specific planning, authoring, and verification tools, we show that modern agents are capable of authoring diverse long-horizon dynamic scenes involving humanoid motion, character animation, and physical dynamics.

\appendix
\clearpage

\section*{Appendix Overview}
\startcontents[appendix]
\printcontents[appendix]{}{1}{\setcounter{tocdepth}{2}}

\newpage

\section{Implementation Details}
\label{sec:supp_implementation}

\subsection{Prompt Rewriting}
\label{sec:rewriting}

\newcommand{\promptheading}[1]{%
  \par\vspace{0.65em}%
  {\color{promptaccent}\bfseries #1}%
  \par\vspace{0.25em}%
}

We use a dedicated prompt rewriting stage to construct the scene and motion descriptions provided to the animation author. The full instructions for the scene and motion prompt rewriters are shown below.

\begin{promptbox}[Scene prompt rewriter]

You are given one animation brief. Write the \textbf{scene entry} for it:
the place the brief happens in. It is staged beside the brief as
\texttt{scene.txt} and is the only description of the place the animation
author gets. A separate pass after yours describes how the performance is
carried out, so leave the action to it: where the brief's bodies shape the
place --- a surface set at their height, a fixture fitted for them --- that is
the place, but how they use it is not. The arrangement is yours, and it must
serve the action: put what the action handles together close together, in the
order the work needs, so consecutive actions are a step or two apart rather
than a walk; the place's other zones and history surround that working core
rather than scattering it. A layout that needs long paths or rapid transit
between distant stations is a design fault.

The author works the animation contract below, which already requires the
whole place at real dimensions, and it reliably builds the room, its
architecture, fixtures, appliances and stored goods without you. What it
reliably leaves out is what a place accumulates, and how its things are made.
That is your job.

\promptheading{Name the default}

Start by naming the default scene: what an agent reasoning briefly builds for
this brief --- the props the action touches, new and unused, evenly spaced on
axis-aligned surfaces, with fixtures around them and uniform light. Say it in
a sentence or two at the start of the entry. Then picture what it is missing,
through one organizing idea.

\textbf{Choose one non-obvious idea whose axis is the place's use over time}:
how it came to be as it is. An idea that explains only the layout gives a
layout; one that explains the place's history generates the layout, its
leftovers, its neighbours and its light together, and they agree because they
come from one account. An entry with three ideas has none.

\promptheading{What the place is missing}

Picture the place as that idea makes it, and write what the default lacks:

\begin{itemize}
    \item \textbf{The other activities it supports}, each with its own zone,
    surfaces and leftovers.

    \item \textbf{What lies beyond each boundary}: what each opening looks
    onto, which also decides the light arriving and what stands behind the
    subject.

    \item \textbf{What earlier use has left}: things put down and not put
    away, part-used, or belonging to someone not in the film.
\end{itemize}

\promptheading{What the objects are missing}

Walk through the objects the same way. The default object is a primitive:
a box with a handle, a cylinder for a pot, a slab for a door. For the objects
the place is recognized by and the ones the action handles, fixtures and
appliances included, picture how the real one is made and write what shows it:

\begin{itemize}
    \item the process that formed it and what that leaves in its shape ---
    pressed, cast, turned, moulded, bent sheet, welded tube, frame and panel,
    carved;

    \item its edges and transitions --- radii, rolled rims, chamfers, lips,
    returns;

    \item its parts and joins --- seams, reveals, fasteners, and hinges and
    handles as pieces of their own;

    \item its thickness where an edge shows, and its secondary structure ---
    ribs, feet, bosses, stiffeners.
\end{itemize}

Describe the form, not how to model it. Things that are separate in reality
are separate objects: what is carried, fitted, stored in or removed from
something else is not fused to it. Wear and thin deposits belong in the
material, not the geometry.

\promptheading{Rules}

\textbf{State the extent.}
Give a plausible extent, or the dimension that decides it; left free, a place
comes out at an arbitrary size and reads as a sparse hall.

\textbf{Place by relation.}
Write a placement that matters as a relation another surface can be checked
against --- \emph{on edge in the gap between the appliance's side panel and
the wall, faces against both} --- not by name.

\textbf{Buildable and honest.}
Ask only for what the SDK reference below can build. Where the brief and the
place cannot both be true, name the conflict, say what decides it, and require
the author to state its choice.

\promptheading{Writing it}

Dense prose. Name a number only where something turns on it; dimensions,
palette, positions and counts are otherwise the author's. Write in the register
of someone who knows the subject: specific nouns, no hedging, no adjectives
doing the work of facts. Describe the place, not your method or the author's
process.

\end{promptbox}

\begin{promptbox}[Motion prompt rewriter]

You are given an animation brief and the scene entry written for it. Write the
\textbf{motion entry}: how the performance in that brief is physically carried
out. It is handed to the animation author with the brief and the scene entry,
before the author plans its states and shots.

The author works the animation contract below, whose stage 1 already asks for
an ordered action series with preconditions, results and changes of control.
Do not restate the brief as a list of actions or repeat the scene entry's
arrangement or its conclusions; the author has both, so build on the scene
entry's numbers and decisions rather than restating them. Your job is what an
author otherwise discovers only after committing to a plan: how the action
actually works.

When the brief involves a person, humanoid robot or animal, read the staged
\texttt{humanoid\_motion} skill before writing and describe a performance that
follows it.

\promptheading{Name the default}

Start by naming the default performance: what an agent reasoning briefly would
animate for this brief. That is usually one pose at each end of each verb,
objects that follow an effector with nothing holding them, bodies that glide
between work positions, and everything still except the one thing acting.
Name it in a sentence or two at the start of the entry, then use the three
lenses below to find what it is missing.

\promptheading{Technique}

A brief names outcomes; the bodies reach them through a specific sequence of
contacts. For each consequential action, how would someone who does this often
do it with \emph{these} bodies, joints and tools? How is each thing held, what
serves as a support or a third hand, and where does a mechanism's energy come
from and how does each stage pass it to the next? Derive the technique from the
capabilities you are given --- the preloaded assets' joints and limits, the
scene's heights and extents --- and compute what can be computed. Where the
brief asks for something a body cannot do, name the conflict, give the
technique that best keeps what the brief is about, and require the author to
state its choice. Where something goes wrong in the brief, give it a physical
cause in the configuration before it, and make the recovery a different
technique rather than a repeat.

\promptheading{Transitions between states}

A state begins wherever what holds, supports or drives a participant changes,
not wherever the brief has a verb. Every movable thing is always held, resting,
or moving under a cause you can name. Find the states the brief implies between
its verbs: regrasps, intermediate placements, changes of stance, moments where
two holders share a load. For each transition, say what drives it, what must
stay fixed while it happens, and the order of overlapping changes --- a new
holder takes the load before the old one lets go. Prefer the technique and
order with the fewest state changes and the shortest paths that keep what the
brief is about; if the arrangement forces long or rapid travel, state it as a
requirement on the place. Carry state across the whole: what was opened stays
open, what was set down stays put, and participants not acting are still doing
something.

\promptheading{Spatial relationships}

For each state, give the spatial relationships it needs, including the ones the
brief implies but does not state: which surfaces touch and which bears on
which, where each body stands and faces, reach and clearance for the next
motion, room for the path. Where one part drives, releases or stops another,
name the surfaces that meet. Write the relationships the action depends on as
something an inspection could check at a named state --- a contact, distance,
clearance or joint value --- and prefer what must be present over what must be
absent. Where the performance needs something the scene entry does not provide,
state it as a requirement on the place.

\promptheading{Writing it}

Prose, in the order the performance happens. Name a number only where the action
turns on it. Everything must be authorable: rigid parts on joint trees, with
couplings and deformation coordinated in Blender, so describe a break or a pour
as the pieces or deformation it produces. Leave shots, framing, timing and
style to the author. Write in the register of someone who does this work. Write
the performance as it happens --- \emph{the left gripper clamps the rim while
the right\ldots} --- not as instructions; address the author only for a choice
it must state.
\end{promptbox}

\newpage

\subsection{Character Assets and Motion Sources}
\label{sec:supp_characters}

Characters are the hardest participants for an agent to produce from scratch: modelling and rigging a person or an animal competes with the animation itself for the agent's effort, and hand-keyed locomotion tends to come out stiff.
\method therefore gives the agent a large library of rigged characters to retrieve from, and a basis for their motion.
\paragraph{Character library.}
We build the library from the animated subsets of Objaverse~\citep{deitke2023objaverse} and TexVerse~\citep{zhang2025texverse}: 106{,}976 records (86{,}942 distinct shapes, of which 62{,}205 carry usable motion). Every asset is rendered (textured and untextured views, plus an eight-frame sheet per distinct motion clip) and described by a vision-language model in a structured \emph{asset card}: subject, parts, character traits, visual fidelity, visible defects and the
action of each clip. For the 65{,}430 skinned assets we recover the original file as a native Blender rig and record its skeleton, skin binding and clips.
The importer recognizes the common skeleton
naming conventions (e.g.\ Mixamo, Character Creator, Rigify, Unity/VRM), so that 68\% of the 45{,}692 character rigs map onto a known skeleton family.
Imported characters are frozen native Blender rigs (armature, skinned mesh and materials) at real-world scale, facing a common direction with their feet on the ground, and the agent animates the rig directly rather than a proxy.
\paragraph{Retrieval.}
The prompt names who the characters are (e.g.\ ``the dog is a German shepherd'') rather than where to find them. The agent searches the card database with hard filters and keyword search over the cards' text and clip labels, inspects the hits as summaries, a contact sheet and fixed-scale line-ups, and stages the character it chooses. Staging verifies the bundle's content hash and the digest of its rig file, so the scene is built around exactly the body that is animated.
\paragraph{Motion capture reference.}
Each run receives two read-only motion-capture libraries, each with a searchable text index, from which the agent selects its own reference clips. It searches the indices for the actions its performance needs, inspects the candidate clips, and decides how each one is used: retargeted onto the character and edited into the performance, spliced with other spans, or consulted only for timing, stride and weight transfer while the motion is authored by hand. Nothing selects clips for the agent or requires a clip to be used; the agent states in its notes which motions it based on capture and which it authored itself. The libraries are:
\begin{itemize}
  \item \textbf{Human motion:} the CMU Graphics Lab Motion Capture Database~\citep{cmu_mocap}, 2{,}548 clips from 113 subjects at 120\,fps, covering locomotion, turning, sitting and standing, bending, reaching, carrying, gestures and sports. Its index carries the database's own description of each clip.
  \item \textbf{Dog motion:} the captures of ~\citet{zhang2018mode}, 51 clips (41 minutes) of one dog at 60\,fps. Its index divides them into 1{,}065 labelled spans (walk, trot, canter, stand, sit, lie, sniff, turn in place, jump, and transitions between them), with speeds.
\end{itemize}
Optional retargeting helpers handle the mechanical part when the agent chooses to retarget. For humanoid rigs, the helper maps the source skeleton onto the rig's bone-naming convention, compensates for rest-pose and bone-axis differences, scales root travel by leg length, and grounds and locks planted feet. For rigify quadruped rigs, the helper retargets edited and spliced spans along a path and over uneven ground. Both report foot slip and ground contact for the agent to inspect. Neither helper poses hands, fingers or grips, since the captures record none, and the agent may edit or replace anything the
helpers produce.
\clearpage
\section{Additional Results}
\label{sec:supp_results}

\subsection{More Qualitative Results}
\label{sec:supp_more_results}

\newcommand{\benchmarkpanel}[2]{%
  \par{\centering
  \begin{minipage}{0.98\linewidth}
    \qualrow{#1}{#2}
  \end{minipage}
  \par}\medskip
}

\setcounter{figure}{0}
\renewcommand{\thefigure}{S\arabic{figure}}
\renewcommand{\theHfigure}{S\arabic{figure}}
\captionof{figure}{Results for all 24 prompts in \benchmark.}
\label{fig:additional_results}

\benchmarkpanel{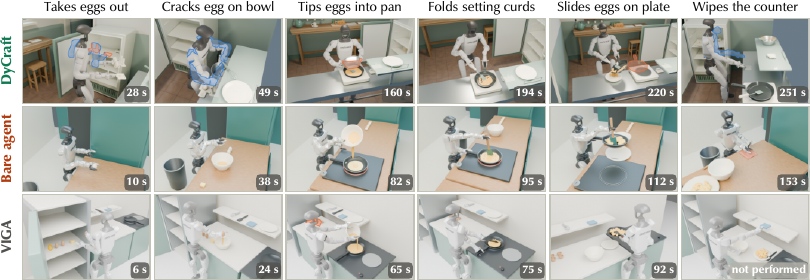}{G1 cooks scrambled eggs, from the fridge to the plate}
\benchmarkpanel{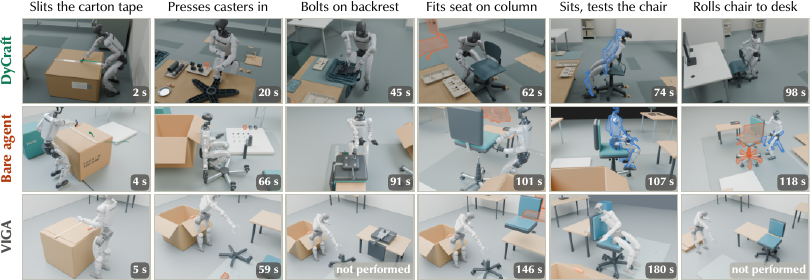}{G1 assembles a flat-pack office chair and rolls it to a desk}
\benchmarkpanel{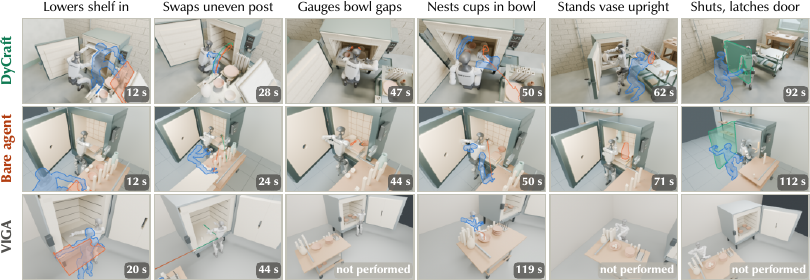}{G1 loads a pottery kiln, replacing an uneven support post}
\benchmarkpanel{Figures/Results/b04_table.pdf}{G1 mops a break room, working backward toward the exit}
\benchmarkpanel{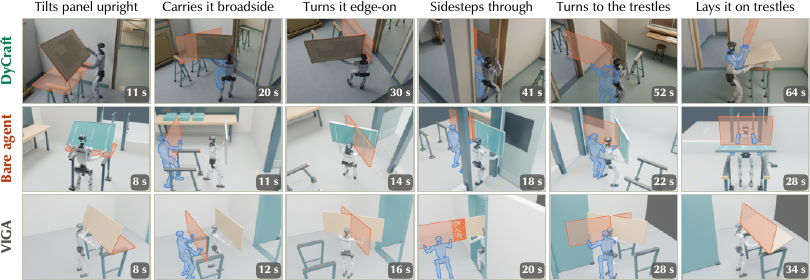}{G1 carries a wide panel edge-on through a narrow doorway}
\benchmarkpanel{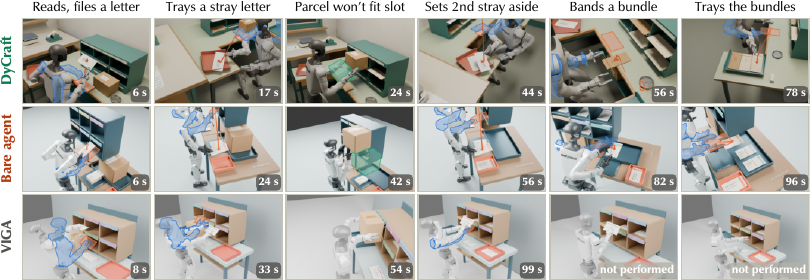}{G1 sorts mail, setting aside misdirected items and an oversized parcel}
\benchmarkpanel{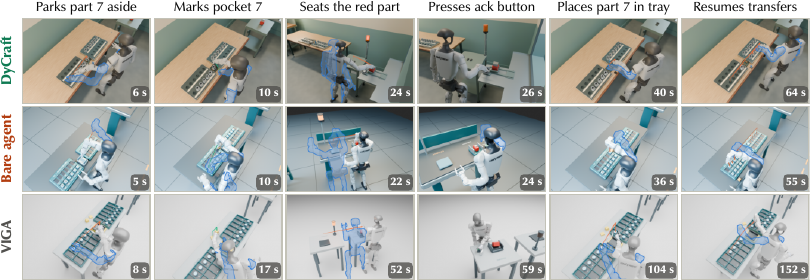}{G1 pauses a tray transfer for an urgent job, then resumes it}
\benchmarkpanel{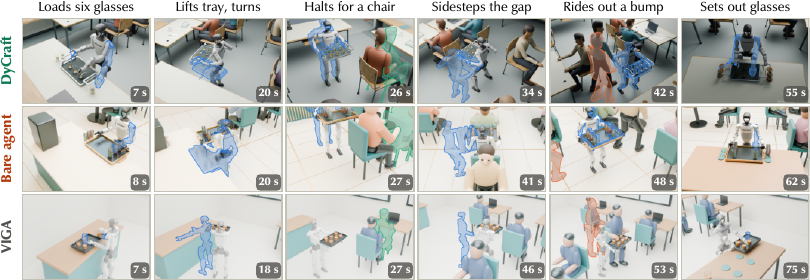}{G1 carries a loaded tray through a crowded room}
\benchmarkpanel{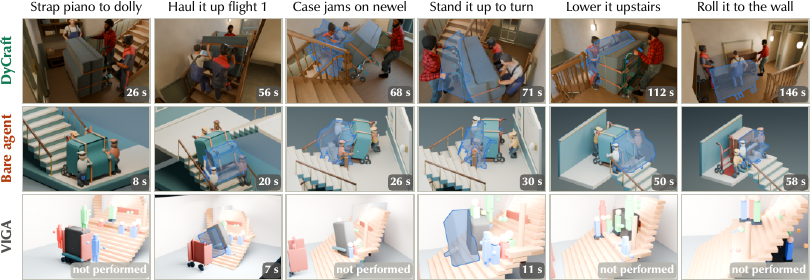}{Three movers carry an upright piano up a stairwell with a tight turn}
\benchmarkpanel{Figures/Results/b10_table.pdf}{Barista serves a takeaway coffee}
\benchmarkpanel{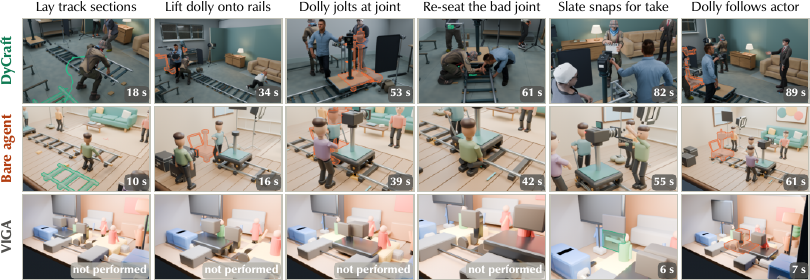}{A film crew sets up and shoots a dolly shot}
\benchmarkpanel{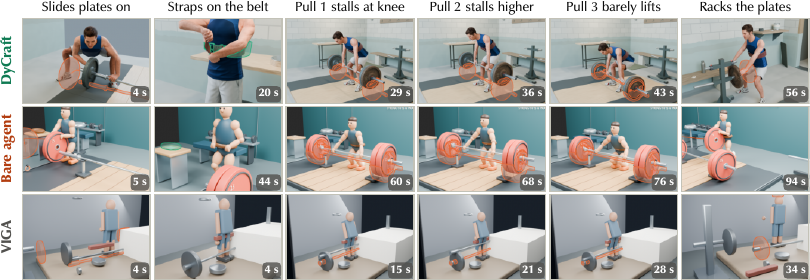}{A weightlifter fails a heavy deadlift and puts the equipment away}
\benchmarkpanel{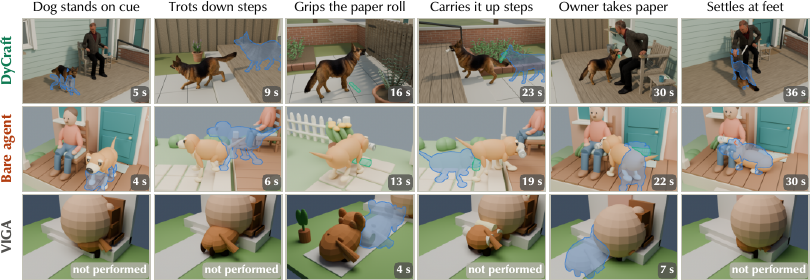}{A dog fetches a newspaper from the garden and delivers it to its owner}
\benchmarkpanel{Figures/Results/b14_table.pdf}{Crow opens a puzzle feeder}
\benchmarkpanel{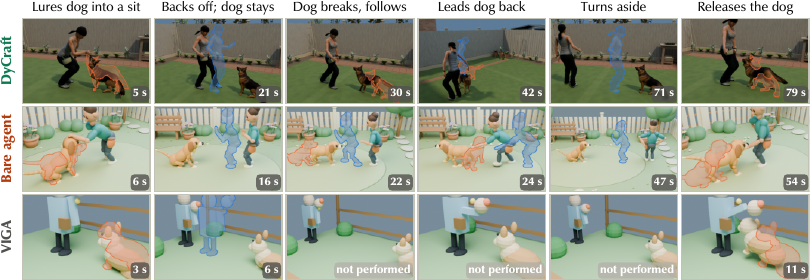}{A trainer teaches a dog to sit and stay through repeated attempts}
\benchmarkpanel{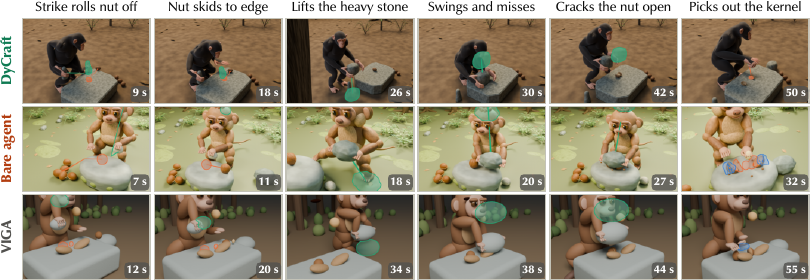}{A young macaque tries different stones to crack a nut}
\benchmarkpanel{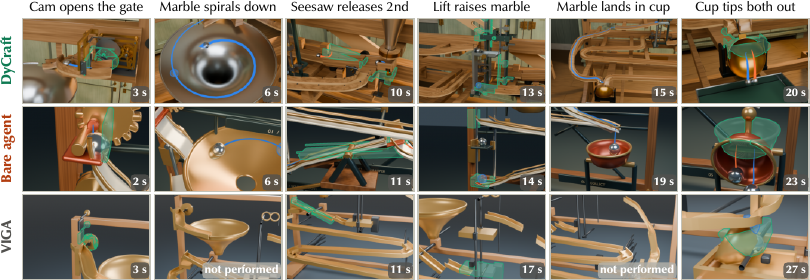}{Marble course with a counterweight lift}
\benchmarkpanel{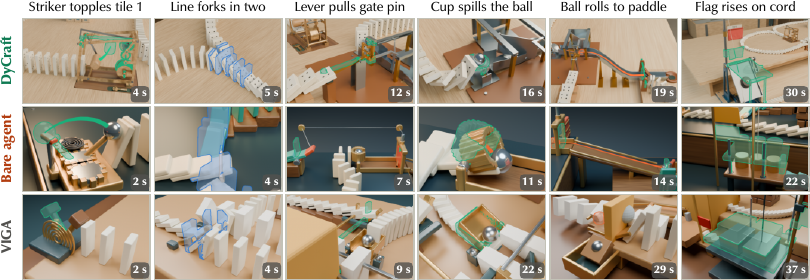}{A branching domino circuit releases a ball through a gate and raises a flag}
\benchmarkpanel{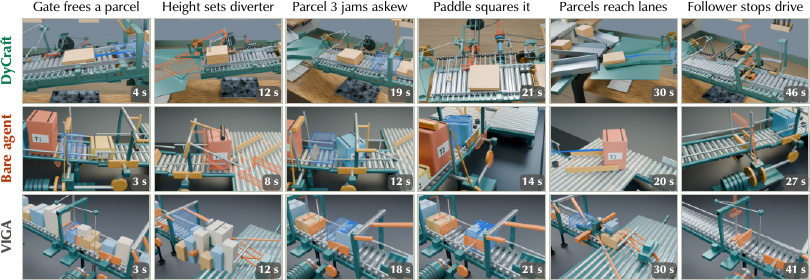}{A gravity-fed machine sorts six parcels by height into three lanes}
\benchmarkpanel{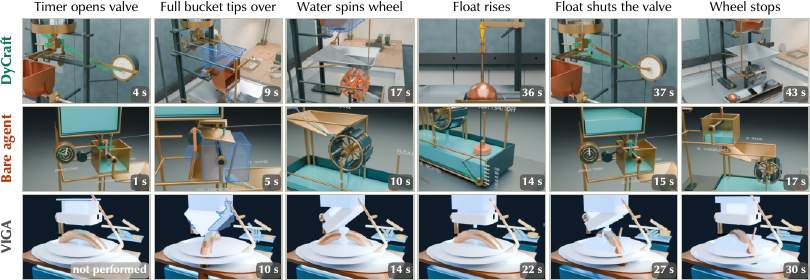}{A tipping bucket transfers water into a tank until a float closes the valve}
\benchmarkpanel{Figures/Results/b21_table.pdf}{An automatic line fills, caps, and labels six bottles}
\benchmarkpanel{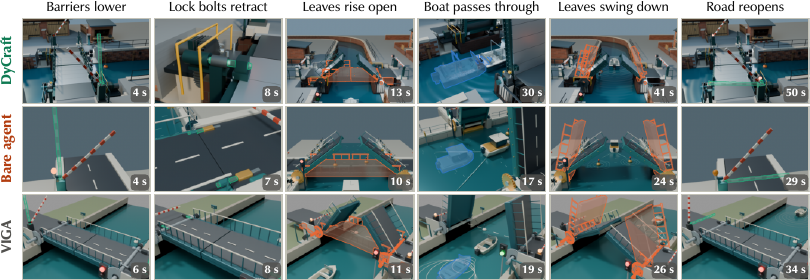}{A drawbridge opens for a passing boat and closes behind it}
\benchmarkpanel{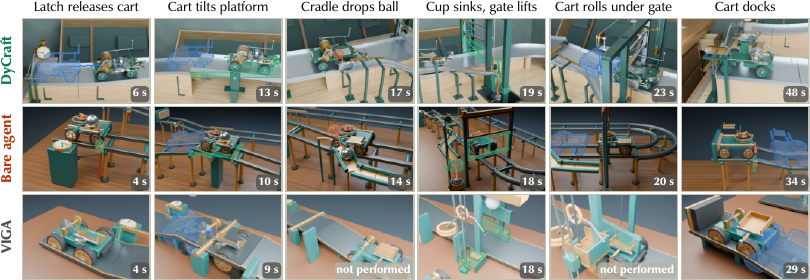}{A spring-powered cart releases a ball to open a gate and completes an obstacle course}
\benchmarkpanel{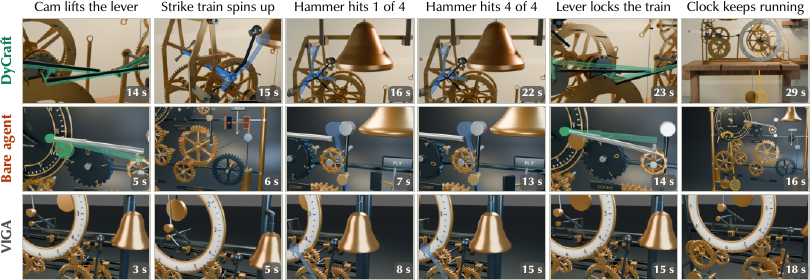}{A weight-driven clock runs and strikes four times}

\clearpage
\subsection{User Study}
\label{sec:supp_user}

In total, 109 individuals participated in the study, and 1,538 binary-choice judgments were recorded. Each participant was assigned five film pairs, with three questions per pair. To ensure response quality, we discarded judgments on pairs where either film played for less than 1 second.

\definecolor{prefBlue}{HTML}{1C5CB5}
\definecolor{prefRed}{HTML}{DC5B56}

\newcommand{\prefcell}[1]{%
  \ifnum#1<40
    \cellcolor{prefRed!55}#1%
  \else\ifnum#1<50
    \cellcolor{prefRed!25}#1%
  \else\ifnum#1<60
    \cellcolor{gray!12}#1%
  \else\ifnum#1<80
    \cellcolor{prefBlue!30}#1%
  \else\ifnum#1<90
    \cellcolor{prefBlue!60}#1%
  \else
    \cellcolor{prefBlue!90}\textcolor{white}{\bfseries #1}%
  \fi\fi\fi\fi\fi
}

\begin{table}[h]
\centering
\caption{\textbf{Detailed human preference results.}
Percentage of binary-choice judgments favoring \method over each baseline
for each prompt in \benchmark. The three aspects are overall 4D realism,
physical plausibility, and motion quality. We recommend viewing these results alongside Fig.~\ref{fig:additional_results} and the videos.}
\label{tab:user_study_detailed}
\small
\setlength{\tabcolsep}{3pt}
\resizebox{\linewidth}{!}{%
\begin{tabular}{l*{6}{>{\centering\arraybackslash}p{1.25cm}}}
\toprule
& \multicolumn{3}{c}{\textbf{Ours vs.\ VIGA (\%)}} &
    \multicolumn{3}{c}{\textbf{Ours vs.\ Bare Agent (\%)}} \\
\cmidrule(lr){2-4}\cmidrule(l){5-7}
\textbf{Prompt} &
\textbf{4D} & \textbf{Physical} & \textbf{Motion} &
\textbf{4D} & \textbf{Physical} & \textbf{Motion} \\
\midrule
A humanoid cooks scrambled eggs, fridge to plate            & \prefcell{91} & \prefcell{82} & \prefcell{91} & \prefcell{78} & \prefcell{89} & \prefcell{78} \\
Flat-pack office chair assembly                             & \prefcell{100} & \prefcell{100} & \prefcell{100} & \prefcell{91} & \prefcell{100} & \prefcell{100} \\
Loading a pottery kiln                                      & \prefcell{92} & \prefcell{91} & \prefcell{91} & \prefcell{80} & \prefcell{80} & \prefcell{60} \\
Mopping a room toward the exit                              & \prefcell{91} & \prefcell{91} & \prefcell{91} & \prefcell{100} & \prefcell{91} & \prefcell{82} \\
Carrying a wide panel through a doorway                     & \prefcell{90} & \prefcell{90} & \prefcell{90} & \prefcell{82} & \prefcell{73} & \prefcell{64} \\
Sorting mail with exceptions                                & \prefcell{100} & \prefcell{100} & \prefcell{80} & \prefcell{91} & \prefcell{73} & \prefcell{73} \\
Handling an urgent job and resuming a tray transfer         & \prefcell{100} & \prefcell{100} & \prefcell{100} & \prefcell{60} & \prefcell{60} & \prefcell{50} \\
Carrying a loaded tray through a crowded room               & \prefcell{91} & \prefcell{82} & \prefcell{82} & \prefcell{90} & \prefcell{100} & \prefcell{90} \\
Moving an upright piano up a stairwell                      & \prefcell{100} & \prefcell{100} & \prefcell{100} & \prefcell{90} & \prefcell{80} & \prefcell{90} \\
A coffee order from arrival to handoff                      & \prefcell{100} & \prefcell{100} & \prefcell{100} & \prefcell{88} & \prefcell{75} & \prefcell{75} \\
A film crew prepares and executes a shot                    & \prefcell{100} & \prefcell{91} & \prefcell{82} & \prefcell{70} & \prefcell{60} & \prefcell{60} \\
An unsuccessful lift followed by putting the equipment away & \prefcell{100} & \prefcell{100} & \prefcell{100} & \prefcell{80} & \prefcell{100} & \prefcell{90} \\
A dog fetches the newspaper                                 & \prefcell{100} & \prefcell{91} & \prefcell{82} & \prefcell{56} & \prefcell{44} & \prefcell{67} \\
A crow opens a puzzle feeder                                & \prefcell{100} & \prefcell{100} & \prefcell{100} & \prefcell{78} & \prefcell{89} & \prefcell{78} \\
Teaching a dog to sit and stay                              & \prefcell{100} & \prefcell{91} & \prefcell{100} & \prefcell{70} & \prefcell{80} & \prefcell{70} \\
A young macaque learns to crack a nut                       & \prefcell{100} & \prefcell{100} & \prefcell{100} & \prefcell{80} & \prefcell{60} & \prefcell{70} \\
Marble route with a counterweight lift                      & \prefcell{82} & \prefcell{91} & \prefcell{82} & \prefcell{40} & \prefcell{40} & \prefcell{44} \\
Branching domino circuit                                    & \prefcell{64} & \prefcell{82} & \prefcell{73} & \prefcell{40} & \prefcell{30} & \prefcell{40} \\
Parcel sorting machine                                      & \prefcell{82} & \prefcell{73} & \prefcell{82} & \prefcell{91} & \prefcell{91} & \prefcell{82} \\
Water tank mechanism                                        & \prefcell{100} & \prefcell{100} & \prefcell{100} & \prefcell{89} & \prefcell{67} & \prefcell{89} \\
Bottle filling and capping line                             & \prefcell{100} & \prefcell{100} & \prefcell{100} & \prefcell{82} & \prefcell{80} & \prefcell{78} \\
Automated drawbridge                                        & \prefcell{73} & \prefcell{45} & \prefcell{55} & \prefcell{90} & \prefcell{60} & \prefcell{90} \\
Spring-powered cart obstacle course                         & \prefcell{100} & \prefcell{91} & \prefcell{100} & \prefcell{55} & \prefcell{36} & \prefcell{36} \\
Weight-driven clock                                         & \prefcell{55} & \prefcell{36} & \prefcell{45} & \prefcell{45} & \prefcell{27} & \prefcell{50} \\
\midrule
\textbf{All tasks} &
\prefcell{92} & \prefcell{89} & \prefcell{89} &
\prefcell{76} & \prefcell{70} & \prefcell{71} \\
\bottomrule
\end{tabular}}
\end{table}

\clearpage
\section{DyBench Benchmark}
\label{sec:supp_dybench}

\subsection{Overview}
\begin{figure}[h]
\centering
\includegraphics[width=\linewidth]{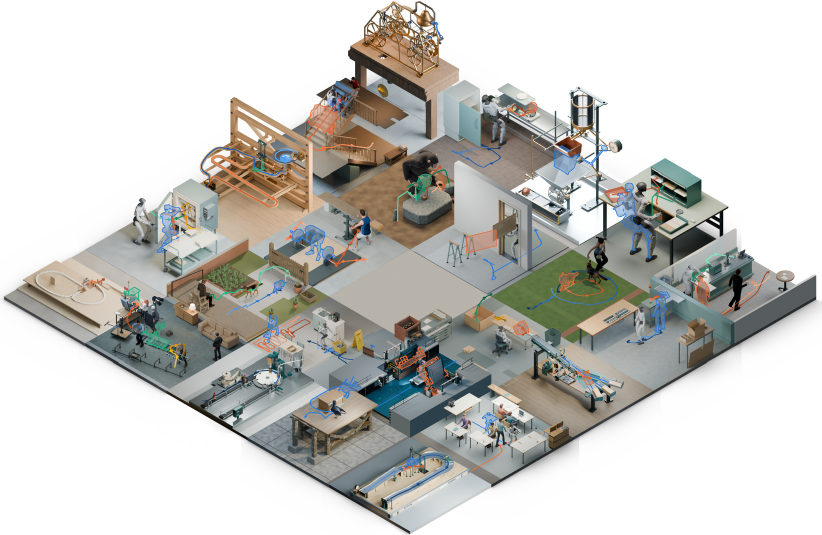}
\caption{Overview of DyBench. The 24 DyBench scenes generated by DyCraft,
arranged on a $5\times5$ board. Colored curves show the trajectories of key
moving objects in each scene.}
\label{fig:dybench_overview}
\end{figure}

DyBench consists of 24 text prompts for long-horizon, multi-shot 4D scene
generation. The prompts vary along two axes:
(1) the \emph{embodiment}, i.e., the primary actor that drives the motion; and
(2) the \emph{organizing idea} of the motion, i.e., the structure that links
the events of the animation (e.g., a multi-step procedure, a causal chain,
recovery from a failed attempt, or an interruption that must be handled).
The benchmark contains 8 tasks driven by a humanoid robot (Unitree G1),
8 tasks driven by human or animal characters, and 8 tasks driven by physical
mechanisms.
Fig.~\ref{fig:dybench_overview} shows all 24 scenes, and
Sec.~\ref{sec:supp_dybench_prompts} lists the full prompts.

\subsection{Full Prompts}
\label{sec:supp_dybench_prompts}

We list the verbatim user prompt for each task below. All methods receive
the same task prompt. Bare agents additionally receive the quality and delivery
guidance described in Sec.~\ref{sec:supp_bare_guidance}.

\begin{dybenchprompt}{A humanoid cooks scrambled eggs, fridge to plate}
A Unitree G1 humanoid robot cooks scrambled eggs by itself in a compact kitchen. Use the existing Unitree G1 robot asset at \texttt{<PATH\_TO\_G1\_ASSET>} as the character; load and reuse its model and articulation rather than creating a new robot. The preparation counter starts clear; three eggs and a portion of butter sit in the refrigerator, with a bowl, whisk, spatula, plate, and cloth on a nearby shelf and a frying pan on the stovetop. An open waste bin stands beside the counter. The robot opens the refrigerator, carries the eggs and butter to the counter, retrieves the bowl, and cracks each egg over it, separating the shells and setting the empty halves aside. It steadies the bowl with one hand and whisks with the other, then centers the pan on the burner, turns on the heat, and adds the butter. Once the butter melts, it lifts and tilts the bowl to pour the eggs into the pan, sets the emptied bowl down, and scrapes and folds the mixture with the spatula as it thickens into soft curds. It turns off the burner, brings over the plate, and tilts the pan while guiding the eggs onto it, then returns the pan to the stove. Finally, it drops the shells into the bin, wipes the preparation counter, puts down the cloth, and closes the refrigerator.
\end{dybenchprompt}

\begin{dybenchprompt}{Flat-pack office chair assembly}
A Unitree G1 humanoid robot assembles a flat-pack office chair by itself in an open-plan office bay. Use the existing Unitree G1 robot asset at \texttt{<PATH\_TO\_G1\_ASSET>} as the character; load and reuse its model and articulation rather than creating a new robot. A sealed carton stands on the floor; the robot cuts its tape with a safety cutter, folds the four top flaps open, and lays out a five-arm star base, five casters, a gas-lift column with its plastic sleeve, the moulded seat, the backrest and its mounting spine, bolts and washers, bolt cover caps, and an Allen key. It presses each caster into a socket of the upturned base, rights the base, slides the sleeve over the column, and inserts the column into the base hub. It places the seat upside down on a low workbench, bolts the backrest spine to its underside, tightens the bolts, and fits the cover caps. It turns the upper assembly upright, lifts it over the column, and lowers it onto the taper. The robot sits briefly to seat the connection, tests the swivel, height lever, and backrest tilt, then stands, puts away the tools, flattens the empty carton beside the recycling bin, and rolls the finished chair to a desk.
\end{dybenchprompt}

\begin{dybenchprompt}{Loading a pottery kiln}
A Unitree G1 humanoid robot loads mixed pottery into a front-loading kiln in a pottery studio. Use the existing Unitree G1 robot asset at \texttt{<PATH\_TO\_G1\_ASSET>} as the character; load and reuse its model and articulation rather than creating a new robot. The kiln is cold and its door is open. A nearby trolley holds bowls, cups, a tall vase, plates, kiln shelves, support posts of several heights, a clearance gauge, and a cone pack. The robot places three posts inside the kiln and lowers the first shelf onto them. It presses lightly at the shelf edges, notices a wobble, lifts the shelf, replaces the uneven post, and checks it again. It loads four bowls with gaps between them, checks the gaps with the gauge, and nests small unglazed cups inside a larger unglazed bowl. It adds taller posts and a second shelf, checks the vase against the available headroom, and places it upright. It loads plates into a rack, adds a third shelf for the remaining small pieces, and places the cone pack where it will be visible through the viewing port. Finally, it checks the clearances around the loaded shelves, withdraws its hands, closes and latches the kiln door, and returns the unused posts and gauge to the trolley.
\end{dybenchprompt}

\begin{dybenchprompt}{Mopping a room toward the exit}
A Unitree G1 humanoid robot mops a tiled office break room, working backward from the far corner toward the doorway. Use the existing Unitree G1 robot asset at \texttt{<PATH\_TO\_G1\_ASSET>} as the character; load and reuse its model and articulation rather than creating a new robot. Chairs are tucked beneath the tables, leaving an open floor with several dusty footprints and a small coffee stain. A mop, a wheeled bucket with a wringer, and a folding wet-floor sign stand in the dry hallway. The robot places the sign beside the entrance, rolls the bucket into the room, dips and wrings the mop, and begins in the far corner. It sweeps the mop in overlapping figure-eight strokes, then steps backward onto the remaining dry floor. After several passes, it rinses and wrings the mop and pulls the bucket closer along the dry strip. It works around the table legs, scrubs the coffee stain with shorter strokes, and continues toward the exit as the cleaned wet area expands. Near the doorway, it moves the bucket into the hallway, mops the final strip, lifts the mop clear, and steps out. It secures the mop in the bucket and leaves the warning sign beside the freshly cleaned room.
\end{dybenchprompt}

\begin{dybenchprompt}{Carrying a wide panel through a doorway}
A Unitree G1 humanoid robot carries a wide rectangular panel from a workshop into an adjoining room through a narrow doorway. Use the existing Unitree G1 robot asset at \texttt{<PATH\_TO\_G1\_ASSET>} as the character; load and reuse its model and articulation rather than creating a new robot. A lightweight panel rests horizontally on two waist-high trestles. It is too wide to pass through the doorway broadside but can fit when turned edge-on. A second pair of trestles stands in the destination room, and the door is secured open. The robot approaches the panel, grips its two accessible edges, lifts it clear of the supports, and rotates it into an upright carrying position. It adjusts its stance and walks toward the doorway with short steps, stopping before the frame to turn the panel edge-on. It guides the leading edge through, sidesteps across the threshold while turning its torso, and brings the trailing edge clear of the frame. Once inside, it turns toward the empty trestles, pauses to align the panel, rotates it back to horizontal, and lowers it onto both supports. It releases its grips, steps back, and checks the panel's placement before returning to a relaxed standing posture.
\end{dybenchprompt}

\begin{dybenchprompt}{Sorting mail with exceptions}
A Unitree G1 humanoid robot sorts a mixed bundle of mail at a mailroom sorting station. Use the existing Unitree G1 robot asset at \texttt{<PATH\_TO\_G1\_ASSET>} as the character; load and reuse its model and articulation rather than creating a new robot. A counter holds a stack of envelopes addressed to six clearly labelled departments, an exception tray, rubber bands, and one small parcel. A bank of matching pigeonholes stands within reach, with a parcel shelf below. The robot squares the bundle, picks up the first envelope, turns its address toward its head, and places it into the corresponding pigeonhole. It continues through several envelopes at a steady pace until it finds one addressed to another building, pauses to inspect it, and sets it in the exception tray. It resumes sorting, then reaches the parcel and checks it against the intended pigeonhole. Finding it too large, it withdraws the parcel and places it on the lower shelf. It works through the remaining envelopes, sets aside a second misdirected item, and checks the now-empty input area. Finally, it gathers the mail from three occupied pigeonholes into separate bundles, secures each with a rubber band, and places them in a collection tray, leaving the exceptions separate.
\end{dybenchprompt}

\begin{dybenchprompt}{Handling an urgent job and resuming a tray transfer}
A Unitree G1 humanoid robot transfers parts between two indexed trays at a workshop bench, interrupts its work for an urgent delivery, and then resumes where it stopped. Use the existing Unitree G1 robot asset at \texttt{<PATH\_TO\_G1\_ASSET>} as the character; load and reuse its model and articulation rather than creating a new robot. The trays each have twelve numbered pockets arranged in two rows. At the start, parts one through six have been transferred to the destination tray, and the robot is lifting part seven from the source tray. A padded temporary holder sits between the trays, and a separate urgent-job station stands a few steps away. An amber lamp illuminates at that station while the robot holds part seven. It pauses, lowers the part into the temporary holder, and places a small marker beside destination pocket seven. It turns, walks to the urgent station, picks up a red component, carries it to a dispatch cradle, and presses the acknowledgement button. When the lamp goes out, it returns to the original bench, checks the marker, retrieves the parked part, and places it into pocket seven. It removes the marker, transfers parts eight through twelve in order, checks the empty source tray and completed destination tray, and steps back from the bench.
\end{dybenchprompt}

\begin{dybenchprompt}{Carrying a loaded tray through a crowded room}
A Unitree G1 humanoid robot carries a tray of drinks from a service counter to a collection table across a busy office lounge. Use the existing Unitree G1 robot asset at \texttt{<PATH\_TO\_G1\_ASSET>} as the character; load and reuse its model and articulation rather than creating a new robot. Six partially filled glasses stand beside a rigid serving tray on the counter. Several people sit at tables along the route, leaving an aisle with one narrow gap between chairs. The robot places the glasses onto the tray, lifts it with both hands, and turns toward the collection table. It walks between the occupied tables until a seated person pushes a chair backward into its path. The robot stops, lets the glasses settle, and waits while the person notices it and pulls the chair inward. It resumes, turns its body sideways to pass through the narrow gap, and brings the tray through at a steady height. Beyond the gap, a passing person lightly bumps its upper arm. The robot pauses, bends its knees slightly, adjusts its stance, and steadies the tray before continuing. At the collection table, it lowers the tray onto the surface, releases its grip, and arranges the six glasses one by one on coasters. It then picks up the empty tray and returns toward the service counter.
\end{dybenchprompt}

\begin{dybenchprompt}{Moving an upright piano up a stairwell}
Three movers carry an upright piano up two flights of stairs in an apartment building, negotiating a tight turn at the half-landing. The piano stands in the entrance hall beside a stair-climbing dolly, moving blankets, and straps. The movers wrap the case, secure it to the dolly, and tip it backward together, with one mover above guiding and two below supporting the load. They find their balance, approach the first step, and work up the flight one step at a time, pausing midway to adjust their grips. At the half-landing, they begin turning toward the second flight, but the piano's case meets the space occupied by the newel post. They stop, back off slightly, and bring the piano almost upright to make room for the turn. The upper mover changes position while the other two hold the load, then they pivot the piano and reposition the dolly toward the next flight. They resume climbing, reach the upper landing, and lower the piano upright onto the floor. Finally, they release the straps, withdraw the dolly, remove the blankets, and move the piano into its position against the hallway wall.
\end{dybenchprompt}

\begin{dybenchprompt}{A coffee order from arrival to handoff}
A barista serves a customer from ordering to collecting a takeaway coffee in a small café. The counter contains a till, paper cups, a marker, an espresso machine, a grinder, a milk pitcher, lids, sleeves, and a cleaning cloth. The customer approaches the till, orders a milk coffee, and gives their name. The barista marks a cup, repeats the order, and takes payment while the customer moves to the collection end of the counter and looks at their phone. The barista grinds and tamps the coffee, locks the portafilter into the machine, and starts the shot. While it extracts, they pour milk into the pitcher and steam it, then remove the cup and pour in the milk. They press on a lid, fit a sleeve, and call the customer's name. The customer does not respond. The barista glances toward them and calls again, raising the cup slightly. The customer looks up, puts away their phone, and approaches. The barista places the drink within reach, the customer takes it and leaves, and the barista wipes the steam wand and counter before returning the pitcher to its place.
\end{dybenchprompt}

\begin{dybenchprompt}{A film crew prepares and executes a shot}
A small film crew sets up a dolly shot in a furnished studio interior. Equipment cases, three track sections, a camera dolly, a lamp, a flag, stands, sandbags, wedges, and marking tape are arranged along the edge of the set. A grip opens a stand, mounts the flag, and secures the base with a sandbag while a gaffer positions the lamp and adjusts its direction. At the same time, a camera assistant lays the track sections end to end, joins them, and levels the track with wedges. The crew places the dolly on the rails and mounts the camera. The gaffer takes a light reading beside the actor's position and adjusts the lamp while the focus puller measures two distances and tapes marks beside the track. An actor steps onto their floor mark for a rehearsal. The dolly grip pushes the camera through the planned move, the focus puller follows the distance marks, and the crew stops to correct a bump at one track joint. They repeat the move, return to the starting positions, and make a final lighting check. The director calls for quiet, the camera starts, an assistant presents and snaps the slate, and the actor performs as the dolly travels along the track.
\end{dybenchprompt}

\begin{dybenchprompt}{An unsuccessful lift followed by putting the equipment away}
A weightlifter attempts a heavy deadlift in a gym, fails to complete it, and puts the equipment away. An unloaded barbell lies on a lifting platform beside four weight plates, collars, a lifting belt, a chalk bowl, and a plate rack. The lifter loads two plates on each end, secures the collars, tightens their belt, and chalks their hands. They approach the bar, set their feet, bend down, and establish their grip. On the first attempt, they pull the bar to knee height, stall, and lower it back to the platform. They release their grip, stand to take a breath, then reset their feet and try again. The second pull rises slightly higher but still stops before lockout, and they lower it once more. After a longer pause, they change their grip and make a third attempt, this time lifting the bar only a short distance before returning it to the floor. They let go, step back, and loosen their belt. Finally, they remove the collars, unload the plates one at a time onto the rack, wipe the bar, and leave it empty on the platform.
\end{dybenchprompt}

\begin{dybenchprompt}{A dog fetches the newspaper}
A dog retrieves a rolled newspaper from the front garden and delivers it to its owner on the porch. The owner sits beside the front door, and a newspaper secured with a paper band lies near the garden gate at the end of a short path. The dog rests beside the chair until the owner gives a verbal cue and points toward the gate. It raises its head, stands, walks to the porch steps, and trots down onto the path. Near the gate, it slows, lowers its nose, and locates the newspaper beside a plant pot. It grips the middle of the roll in its mouth, lifts it, turns, and trots back toward the house. At the steps, it slows and climbs while carrying the newspaper. On reaching the porch, it pauses to adjust its grip, then approaches the owner and raises its muzzle toward the waiting hand. The owner takes hold of the newspaper before the dog releases it. The dog sits beside the chair, the owner places the newspaper on their lap and strokes its head, and the dog settles at their feet.
\end{dybenchprompt}

\begin{dybenchprompt}{A crow opens a puzzle feeder}
A crow works out how to open a puzzle feeder on a low outdoor table. Food is visible behind a transparent sliding hatch, but a removable peg blocks the hatch from opening. A perch stands beside the table. The crow lands near the feeder, walks around it, and tilts its head to inspect the food through the clear panel. It pecks the panel several times, then catches the edge of the hatch with its bill and pushes it sideways. The hatch moves a little and stops against the peg. The crow steps back, looks along the hatch edge, and pecks at the protruding peg. It grips the peg with its bill and pulls, but loses its grip. It changes position, braces beside the feeder, hooks one foot around the peg, and draws it out of the retaining hole. The peg drops onto the table. The crow returns its bill to the hatch edge and slides the hatch far enough to expose the food. It reaches inside, takes a piece, withdraws its head, and hops onto the nearby perch to eat, leaving the hatch open and the peg on the table.
\end{dybenchprompt}

\begin{dybenchprompt}{Teaching a dog to sit and stay}
A trainer teaches a young dog to sit and stay in a quiet fenced garden. The trainer wears a treat pouch, and the dog begins standing in front of them on a clear patch of grass. The trainer holds a treat above the dog's nose and moves it backward until the dog sits, marks the response with a short word, and gives the treat. They repeat the sit with a raised-hand signal, reward it, and then introduce an open-palm stay signal. The trainer steps backward one pace, pauses, returns to the dog, and rewards it while it remains seated. On the next attempt, the trainer takes two steps back, but the dog stands and follows. The trainer stops, guides the dog back to its original place, and asks it to sit again. They repeat the easier one-step stay and reward it, then try two steps again. This time the dog holds its position until the trainer returns. Finally, the trainer backs away three paces and briefly turns sideways before facing the dog again. The dog remains seated. The trainer returns, gives a release cue, and rewards the dog as it gets up and approaches.
\end{dybenchprompt}

\begin{dybenchprompt}{A young macaque learns to crack a nut}
A young macaque attempts to crack a hard-shelled nut on a flat stone in a forest clearing. Several nuts lie beside the stone, with a small cobble nearby and a larger, heavier stone a short distance away. The macaque picks up a nut, places it on the flat surface, and lifts the small cobble in one hand. It strikes the nut, but the nut rolls off the stone without opening. The macaque follows it, retrieves it, and places it back near the center. A second strike sends the nut skidding to the edge, still intact. It retrieves the nut again, sets down the small cobble, and moves to the heavier stone. It lifts that stone with both hands, brings it over the nut, and strikes, but misses to one side. The macaque lowers the stone, repositions the nut, widens its stance, and adjusts its grip. It raises the stone again and strikes squarely, cracking the shell. It sets the stone down beside the anvil, separates the shell pieces with its fingers, picks out the kernel, and sits beside the scattered fragments to eat.
\end{dybenchprompt}

\begin{dybenchprompt}{A marble route with a counterweight lift}
A mechanical marble course runs on a freestanding wooden frame with ramps, a spiral funnel, a seesaw, a counterweight lift, and a finishing cup. No people, animals, or robot characters appear. At the start, a steel marble rests behind a closed gate at the top, a second marble sits in a pocket on the raised end of the seesaw, and the lift platform waits at the bottom of its guide rails. A clockwork cam opens the starting gate, releasing the first marble down a curved ramp and into the funnel. It circles inward, drops through the center, and lands on the lowered end of the seesaw, tipping it and releasing the second marble onto a separate track. The first marble rolls off the seesaw onto the lift platform and depresses a latch. A suspended counterweight descends, raising the platform until its edge meets an upper exit ramp. The marble rolls out, crosses a short jump onto a catch ramp, and enters the finishing cup. Meanwhile, the second marble follows a longer zigzag route and arrives in the same cup a moment later. The cup tips under their combined weight, drops both marbles into a collection tray, and returns against its stop.
\end{dybenchprompt}

\begin{dybenchprompt}{A branching domino circuit with a mechanical finish}
A branching domino circuit unfolds across a large tabletop, combining falling tiles with levers, sliding gates, and a weighted final platform. No people, animals, or robot characters appear. A wound clockwork striker releases and knocks over the first domino, starting a line that divides into two routes. The left branch climbs a series of shallow steps and ends by pushing a lever that withdraws a locking pin from a sliding gate. The right branch follows a longer curved path around a tower of wooden blocks and ends by tipping a small cup containing a steel ball. The ball rolls toward the gate, reaches it after the pin has been withdrawn, and pushes through onto a descending track. At the bottom, it strikes a paddle connected to a second domino line on a lower level. This line crosses a narrow bridge and knocks a hinged support away from beneath a weighted platform. The platform lowers slowly through a damped linkage, pulling a cord that raises a small flag. The final dominoes settle, the ball comes to rest in a pocket, and the raised flag remains visible above the completed circuit.
\end{dybenchprompt}

\begin{dybenchprompt}{A gravity-fed parcel sorting machine}
A gravity-fed sorting machine processes six rigid parcels from a sloped loading rack into three collection lanes. No people, animals, or robot characters appear. The parcels have equal width but three different heights, with two parcels of each height mixed in the queue. A motor-driven escapement releases one parcel at a time onto a roller track. Each parcel passes beneath a height-sensing lever that positions a downstream diverter through a mechanical linkage, routing short, medium, and tall parcels into separate chutes. After the first two parcels have been sorted, the third turns slightly and catches between the guide rails. The drive continues through a slipping clutch until a cam-operated side paddle pushes the parcel square with the track. It moves forward, passes the height lever, and enters its collection lane. The remaining parcels follow through the same release and sorting cycle. Each lane's end stop holds its two parcels in a short queue. When the loading rack becomes empty, a follower drops into the vacant feed position and disengages the drive clutch, leaving the escapement stopped and all six parcels sorted.
\end{dybenchprompt}

\begin{dybenchprompt}{A water tank fills, tips, and resets}
A tabletop water mechanism transfers water from an elevated reservoir through a tipping bucket into a lower collection tank. No people, animals, or robot characters appear. The reservoir begins full, the bucket rests upright on a pivot, and the collection tank is empty. A spring-driven timer opens the reservoir valve, sending a narrow stream into the bucket. As the water rises, the bucket's changing center of mass carries it past its tipping point, and it rotates forward to discharge into a broad funnel. A small counterweight returns the empty bucket upright, where it begins filling again. Water from the funnel turns a paddle wheel before draining into the lower tank. The bucket completes several fill-and-tip cycles, and the wheel starts with each discharge and slows between them. A float in the collection tank rises as water accumulates. Near the final level, the float pulls a linkage that closes the reservoir valve. The remaining water in the feed channel drains into the bucket, leaving it partly filled below its tipping threshold. The paddle wheel coasts to a stop, and the system settles with the upper valve closed and the lower tank full.
\end{dybenchprompt}

\begin{dybenchprompt}{An automatic bottle filling and capping line}
An automatic production line fills, caps, and labels a batch of six transparent bottles. No people, animals, or robot characters appear. Empty bottles queue on an infeed conveyor beside a liquid reservoir, a cap chute, and a label reel. A rotating star wheel separates the bottles and brings the first beneath a filling nozzle. The bottle stops, the nozzle lowers, and a measured portion of colored liquid enters before the nozzle rises and the star wheel advances. At the next station, a cap drops from the chute into a holder and a rotating head presses and tightens it onto the bottle. The bottle then passes between powered rollers that turn it against a dispensed label, wrapping the label around its body. As more bottles enter, filling, capping, and labelling occur simultaneously at different stations. One label feeds slightly short, causing the label station to pause while a tension arm takes up slack and advances the reel to the next registration mark. Processing resumes, and the last bottles pass through. All six collect behind an outfeed stop, the empty infeed sensor halts the indexing wheel, and the remaining stations finish their cycles before stopping.
\end{dybenchprompt}

\begin{dybenchprompt}{A drawbridge opens for a passing boat}
An automated canal drawbridge opens for an unoccupied motorboat and closes again after it passes. No people, animals, or robot characters appear. The bridge begins horizontal and locked across the canal, with raised road barriers and the boat approaching slowly along the water. As the boat reaches an approach marker, warning lamps flash and the road barriers lower. Once both barriers reach their stops, locking bolts retract from the bridge deck. A geared drive turns the lifting shafts, raising the two bridge leaves while their counterweights descend into side housings. The boat slows near the opening and waits until the leaves reach their fully raised positions and the canal signal changes. It then travels between the piers, leaving a small wake that rocks two moored buoys. After its stern clears the far marker, the drive reverses and lowers both leaves toward the center. They settle onto their supports, the locking bolts slide into place, and the road barriers rise. The warning lamps go dark as the boat continues down the canal and the disturbed water gradually settles.
\end{dybenchprompt}

\begin{dybenchprompt}{A spring-powered cart completes an obstacle route}
A spring-powered cart travels through a mechanical obstacle course on a tabletop track. No people, animals, or robot characters appear. The cart starts behind a retaining latch with its drive spring wound, carrying a steel ball in a shallow cradle. A timer releases the latch, and the cart rolls forward, climbs a gentle ramp, and passes over a hinged platform that tilts under its weight. As it descends the far side, a fixed guide engages a lever on the cart and opens the cradle. The ball drops into a chute and rolls ahead to a locked crossing gate. It lands in a weighted cup, lowering the cup and pulling the gate upward by a cord. The cart reaches the crossing, passes beneath the raised gate, and enters a curved section that turns it back toward the start. It crosses a rough strip that slows its wheels, then climbs a final short incline using the remaining spring energy. At the top, it rolls into a padded docking stop and engages a catch. The drive spring unwinds the last fraction through a slipping clutch, leaving the cart parked and the ball held in the lowered gate cup.
\end{dybenchprompt}

\begin{dybenchprompt}{A weight-driven clock runs and strikes}
A weight-driven clock with exposed gears advances to the hour, strikes four times, and returns to normal running. No people, animals, or robot characters appear. The clock begins a short interval before four o'clock, with its pendulum swinging, its driving weights raised, and the hour-striking mechanism held by a latch. The escapement releases the gear train tooth by tooth as the minute hand approaches twelve. A cam on the motion works gradually lifts the striking release lever. At the hour, the lever clears its stop and releases a separate striking train, whose weight begins descending as a fly governor spins up. A pinned wheel lifts a hammer and lets it fall against a bell four times, with the hammer withdrawing after each impact. A counting mechanism advances with each strike and, after the fourth, allows the locking lever to drop into its stop. The striking train halts and the governor settles, while the main escapement and pendulum continue running. End with the hands showing just after four, the hammer at rest, and the timekeeping gears still advancing.
\end{dybenchprompt}

\clearpage
\section{Bare Agent Guidance}
\label{sec:supp_bare_guidance}

To compare fairly with bare agents, we append the same natural-language
guidance to every DyBench prompt given to the bare-agent baselines. The guidance
sets out the quality and delivery requirements for the final animation
(completeness, physically credible motion and causality, visual finish, and the
required output files). It does not prescribe DyCraft's prompt rewriting,
planning, 3D storyboarding, motion authoring tools, or verifiers, and it leaves
the choice of workflow, methods, and validation to the agent. We reproduce the
guidance verbatim below.

\begin{promptbox}[Animation design and delivery guidance]

Create a complete, polished Blender animation realizing the task above.
Author the scene and animated performance, and deliver the editable project,
complete rendered film, and supporting evidence. Reuse any explicitly supplied
assets as directed, preserving the identity, geometry, and articulation the
task requires.

\textbf{Your decisions.}
You own the creative direction, visual style, designs, staging, choreography,
duration, pacing, frame rate, aspect ratio, cameras, editing, sound, rigging,
animation techniques, renderer, production workflow, and validation methods,
subject to the task and these delivery requirements.

Choose how to allocate your time and compute. No production stages, review
cycles, critics, planning documents, animation algorithms, or shot counts are
prescribed. Choose methods that produce a convincing, editable final result
within the resources actually available.

\textbf{Scope and completeness.}
Show the complete event, including the consequential actions and intermediate
states needed to make the requested outcome understandable and credible.
Essential interactions must occur in the animated scene and be visible in the
finished film. Descriptions, attractive stills, and an already completed result
do not establish that an omitted action occurred.

Develop the setting, subjects, and supporting assets to the extent required by
the task. The scene must feel complete and intentional at establishing, working,
and close inspection distances. Detail should serve the task and the chosen
visual language. Maintain a coherent scene beyond the principal view.

\textbf{Motion and causality.}
Motion must convey credible timing, weight, balance, forces, and material
behavior appropriate to the subject and the task. Articulated mechanisms must
move in ways their construction permits. Contacts, grasps, releases, support
changes, tool engagement, and transfers must be consistent with the action and
the load. Choose the means of achieving this, including authored motion,
simulation, or a combination.

Objects retain their identity and state through time. Changes in position,
attachment, shape, contents, or appearance must have a cause consistent with
the task. There must be no unintended sliding, floating, penetration,
teleportation, popping, duplication, disappearance, or reset. Contents and
materials that are transferred must have credible origins and destinations.
These requirements apply throughout transitions and between key poses, as well
as across edits. State changes and contact must remain consistent with what the
acting subject is actually doing.

\textbf{Visual finish and presentation.}
Achieve a professional level of modeling, materials, lighting, composition, and
animation in the visual style you choose. Important forms, actions, and contacts
must remain readable. The finished sequence must be temporally stable and free
of distracting geometry, shading, rendering, and encoding defects.

Unless the task specifies the timing, choose a duration and presentation that
communicate the complete event with readable timing and an appropriate sense of
effort. The film must contain the full animated performance. Camera placement,
cuts, darkness, motion blur, depth of field, or effects must not conceal missing
actions or unresolved defects. Other relevant viewpoints must reveal a coherent
underlying scene.

\textbf{Required delivery.}
All paths below are relative to your assigned task directory. Use these paths
unless the task explicitly supplies different delivery locations.

\begin{itemize}

  \item \texttt{output/blend/final.blend}: the complete, meaningfully editable
  Blender scene and its actual finished animation. Preserve useful controls,
  mechanisms, object identities, and scene structure so the performance can be
  edited and retimed. Set the scene frame range and frame rate to the intended
  performance.

  \item \texttt{output/film/final.mp4}: one complete film rendered from that
  final scene, encoded as H.264 MP4 at a constant frame rate. Unless the task
  specifies a different resolution, deliver at least 2,073,600 pixels per frame
  (the pixel count of 1920 x 1080); choose the aspect ratio. Choose an encoding
  quality that preserves the required visual detail. The film must have complete
  frame coverage and no missing, black, corrupt, unintentionally duplicated, or
  obsolete frames. Still-image slideshows and viewport captures do not meet this
  requirement.

  \item \texttt{output/evidence/}: enough additional rendered visual evidence
  to assess essential interactions and the underlying scene where the film alone
  is insufficient. Choose its format, viewpoints, and amount. Identify which
  requirements the evidence establishes; no fixed still or alternate-film count
  is required. If the film sufficiently establishes a requirement, reference
  its relevant time interval in the notes instead of duplicating it.

  \item \texttt{output/NOTES.md}: delivered paths; Blender version; exact
  commands and any included scripts needed to reproduce the film from the saved
  scene; enough information to edit its principal controls and timing;
  substantive task assumptions; validation evidence; and any remaining
  limitations.

  \item Include the scripts, textures, dependencies, and persistent caches
  actually needed to edit, reopen, and reproduce the scene and film, using
  portable paths or packed resources. Include the source scripts used to author
  the result if authoring was scripted.

\end{itemize}

The saved scene must reproduce the delivered motion and appearance after a
fresh reopening in the documented environment. It must not depend on unsaved
state, transient authoring-session caches, playback history, or an undocumented
external process. Persist or bake generated motion and simulation results as
needed. Describe any script execution required for reproduction. Save the
project with \texttt{bpy.ops.wm.save\_as\_mainfile}. Renders and evidence must
correspond to the delivered final scene.

\textbf{Environment and boundaries.}
This is a standalone task. Keep authored files, caches, temporary data, and
outputs inside the assigned task directory, except for runtime locations
explicitly provided by the execution environment. Do not read, search, copy, or
modify other solutions, experiment arms, repository content, or unrelated user
files. Use only the supplied task inputs and permitted runtime resources.

Installed tools and their documentation may be used. External factual or API
documentation may be consulted if available tools and permissions permit it.
Do not download an existing scene, character rig, or animation to substitute
for the assignment. Explicitly supplied starting assets are permitted and must
be reused when the task requires them. Do not load or invoke skills,
prior-session memories, or instructions from other experiments.

Use the interpreter, Blender version, hardware, and tool access described by the
execution environment. Choose rendering methods and settings within the
resources actually available.

\textbf{Validation, completion, and available time.}
Completion means satisfying the task and the quality and delivery requirements.
Successful execution, saving, rendering, encoding, or passing a structural
check alone does not establish completion. Establish the quality of the actual
saved result using methods you choose and report the evidence honestly in
\texttt{output/NOTES.md}. Inspect the resulting animation and presentation,
including transitions and essential interactions; isolated successful poses do
not establish that motion between them is correct. Distinguish what you
inspected or tested from what you inferred or left unverified. Do not claim
that uninspected frames, unsatisfied requirements, or unrun checks passed.

Use the wall-clock budget and absolute deadline if supplied with the execution.
Run \texttt{date} to read the current time and pace yourself against that
deadline. The budget includes all authoring, rendering, encoding, and
validation. Work autonomously within it and allocate your own time and compute;
finish earlier when the requirements are met. At the deadline, preserve and
report the actual result with unmet requirements identified. A time limit does
not convert an unfinished artifact into a successful delivery. This guidance
does not itself set or enforce a wall-clock limit.

Before the session ends, write \texttt{context\_summary.md} in the task
directory. Include the final status, delivered scene, film, evidence and
documentation paths, checks actually run and their results, and any unmet
requirements. For unfinished or incorrect work, describe attempted repairs and
a useful next step. Reference files and relevant lines instead of reproducing
the task, complete source files, or tool logs. Write this summary directly; no
skill is required.

\end{promptbox}

\end{document}